\documentclass{article}
\usepackage{ijcai22}

\usepackage{pifont}

\usepackage{graphicx}

\usepackage{mathtools}
\usepackage{multirow}
\usepackage{times}
\usepackage{soul}
\usepackage{url}
\usepackage[hidelinks]{hyperref}
\usepackage[utf8]{inputenc}
\usepackage[small]{caption}
\usepackage{graphicx}
\usepackage{amsmath}
\usepackage{amsfonts}
\usepackage{amsthm}
\usepackage{booktabs}
\usepackage{algorithm}
\usepackage{algorithmic}
\usepackage[export]{adjustbox}
\newtheorem{example}{Example}

\title{Learning First-Order Rules with Differentiable Logic Program Semantics}

\author{
Kun Gao$^1$\and
Katsumi Inoue$^2$\and
Yongzhi Cao$^{1}$\And
Hanpin Wang$^{3,1}$\footnote{Contact Author}\\
\affiliations
$^1$Key Laboratory of High Confidence Software Technologies (MOE), School of Computer Science, Peking University\\
$^2$National Institute of Informatics\\
$^3$School of Computer Science and Cyber Engineering, Guangzhou University\\
\emails
kungao@pku.edu.cn, 
inoue@nii.ac.jp, 
\{caoyz, whpxhy\}@pku.edu.cn
}

\begin{document}

\maketitle

\begin{abstract}

Learning first-order logic programs (LPs) from relational facts which yields intuitive insights into the data is a challenging topic in neuro-symbolic research.
We introduce a novel differentiable inductive logic programming (ILP) model, called differentiable first-order rule learner (DFOL), which finds the correct LPs from relational facts by searching for the interpretable matrix representations of LPs. 
These interpretable matrices are deemed as trainable tensors in neural networks (NNs). The NNs are devised according to the differentiable semantics of LPs.
Specifically, we first adopt a novel propositionalization method that transfers facts to NN-readable vector pairs representing interpretation pairs.
We replace the immediate consequence operator with NN constraint functions consisting of algebraic operations and a sigmoid-like activation function. 
We map the symbolic forward-chained format of LPs into NN constraint functions consisting of operations between subsymbolic vector representations of atoms. 
By applying gradient descent, the trained well parameters of NNs can be decoded into precise symbolic LPs in forward-chained logic format.
We demonstrate that DFOL can perform on several standard ILP datasets, knowledge bases, and probabilistic relation facts and outperform several well-known differentiable ILP models. 
Experimental results indicate that DFOL is a precise, robust, scalable, and computationally cheap differentiable ILP model. 

\end{abstract}

\section{Introduction}\label{sec1}

Nowadays, knowledge discovery is an important technique for people acquiring knowledge from either large or complex realistic datasets. 
Relational data mining constructs a human-readable representation from relational datasets. 
An explicit logic program (LP) can be a clear explanation to complex, incomplete, or noisy relational datasets due to the limitations of current techniques, especially in the datasets from business, biology, and medicine. 
However, learning high accuracy LPs from realistic relational datasets is still a hardcore problem in the field of machine learning. 

Inductive logic programming (ILP) is firstly proposed by \cite{muggleton1991inductive} as a combination of inductive learning and logic programming technique has contained several methods. 
The purely symbolic ILP methods \cite{Muggleton2012} learning LPs typically support lifelong learning and have more explainability \cite{CropperDM20}. 
However, some purely symbolic approaches fail in ambiguous realistic facts.
By employing neural networks (NNs), differentiable ILP models generate explicit logic rules in a fast and precise manner from realistic relational facts. 
However, NN-based ILP models usually require many examples \cite{Dong2019} to learn concepts, while symbolic ILP methods only need a few examples~\cite{CropperDM20}.

In this paper, we propose a differentiable ILP method called differentiable first-order rule learner (DFOL), which generates first-order LPs from positive examples and background assumptions without logic templates. 
In DFOL, a novel propositionalization method transfers relational facts into NN-readable data, i.e., vector pairs representing interpretation pairs.
Besides, the matrix embeddings of LPs are regarded as the trainable parameters in NNs. 
Based on the differentiable immediate consequence operator and the forward-chained format of LPs, we design some differentiable constraint functions to guide the NNs to find the correct embeddings of LPs. 
Finally, DFOL demonstrates a high degree of interpretability as the parameters in NNs can be decoded to human-readable symbolic rules.
Simultaneously, the correct symbolic rules can be encoded to the fixed parameters in NNs to facilitate the training process.

The main contributions are summarized below: (1) 
A flexible, precise, fast, and robust differentiable first-order rule learning model is proposed. (2) The proposed model is interpretable. We can not only extract symbolic LPs from the model but also embed LPs into NNs.  (3) We demonstrate that our method outperforms the baselines on most datasets, including small ILP datasets and large knowledge bases.

\paragraph{Related Work.}
Neuro-symbolic models have been focused on over the past decades.  
For learning propositional LPs, d'Avila Garcez \textit{et al.}~\shortcite{GarcezBG01} and Lehmann \textit{et al.}~\shortcite{LehmannBH10} designed algorithms to extract propositional LPs from NNs.  On the other hand, Gao \textit{et al.}~\shortcite{Gao2021} generated propositional LPs using NNs from input-output pairs of the immediate consequence operator of LPs by making the symbolic
method proposed by Inoue \textit{et al.}~\shortcite{Inoue2014} be differentiable.
However, compared with first-order LPs, propositional LPs have less ability to describe relational facts.
For first-order LPs, the similar work includes~\cite{DBLP:journals/ai/EvansHWKS21,evans2018learning,DBLP:conf/nips/Rocktaschel017,DBLP:journals/jair/SourekAZSK18}. 
In their models, the explicit LPs are learned based on the given templates. 
These models need to learn the weights given rules or fill the predicates in rule templates. 
However, we construct LPs without any explicit templates but follow the forward-chained format. 
Thus, DFOL is able to find the correct rules in a flexible way, which means that we do not need any prior knowledge about the tasks, and the generated LPs may have more diverse forms than those generated from strongly biased templates.

From the perspective of the representations of LPs, Qu \textit{et al.}~\shortcite{qu2020rnnlogic} regarded LPs as latent variables.
They developed an EM-based algorithm for extracting LPs through a rule generator and a reasoning predictor. 
Kaur \textit{et al.}~\shortcite{KaurKJKN19} regarded a rule as a lifted random walk. 
In contrast to them, DFOL uses small matrix embeddings to encode LPs.
Hence, with the help of the differentiable semantics of LPs, NNs can be adopted in DFOL to search the embeddings fastly and robustly.   
When adopting NNs to get LPs,  Yang \textit{et al.}~\shortcite{YangFanetc} used differentiable operations to learn the embeddings of LPs. 
However, the embeddings of LPs in DFOL are more interpretable, and we do not need any algorithms to transfer subsymbolic matrices into symbolic LPs. 
CILP++  proposed by Fran{\c{c}}a \textit{et al.}~\shortcite{Franca2014} uses the bottom clauses propositionalization method and three-layer NNs to imitate the immediate consequence operator of LPs. 
However, CILP++ cannot be interpreted into symbolic LPs directly. 
Similarly, Teru \textit{et al.}~\shortcite{teru2020inductive} and Hohenecker and Lukasiewicz~\shortcite{hohenecker2020ontology} used embedding methods to perform relation prediction tasks, which are also induction tasks but do not generate explicit LPs as the results.
Excepting these induction tasks, d'Avila Garcez and
Zaverucha~\shortcite{DBLP:journals/apin/GarcezZ99} and \text{Serafini} and d'Avila Garcez~\shortcite{DBLP:conf/nesy/SerafiniG16} considered deduction tasks with NNs and pre-defined LPs.

\section{Preliminaries}
\label{Background_knowledge}
We recall the concepts of LPs, ILP, propositionalization methods, and differentiable semantics of LPs in this section.

\subsection{Logic Programs}
A (definite) logic program $P$ consists of several rules, and each rule $r$ is described as: 
$\alpha_h \leftarrow \alpha_1 \land  \alpha_2 \dots \land \alpha_n \ (n\geq 0)$,
where $\alpha_h$ is the head atom denoted as $head(r)$; $\alpha_i$'s $(0\leq i \leq n)$ are the body atoms, 
and the conjunction $\alpha_1 \land  \alpha_2 \dots \land \alpha_n$ is the body of $r$. 
The set of all body atoms of a rule $r$ is denoted as $body(r)$.
A set of rules with the same head atom $\alpha_h$ called a same head logic program (SHLP):
$\alpha_h$ $\leftarrow$ $\beta_1$, $\alpha_h$ $\leftarrow$ $\beta_2$, ..., $\alpha_h$ $\leftarrow$ $\beta_m$, 
can be identified with a rule of the form:
$\alpha_h$ $\leftarrow$ $\beta_1$ $\lor$ $\beta_2$ $\lor$ $\dots$ $\lor$ $\beta_m$,
where each $\beta_i$ is the body of the $i$-th rule and $\beta_1$ $\lor$ $\beta_2$ $\lor$ $\dots$ $\lor$$\beta_m$ is a disjunction of conjunctions of literals, i.e., a disjunction normal form formula.
In first-order LPs, each atom $\alpha$ is a tuple $p(t_1, t_2, \dots,t_n)$, where $p$ indicates $n$-ary predicate and $t_1, t_2 \dots, t_n$ are terms, and a term is either a variable or a constant. 
When an atom has no variable, the atom is a ground atom.
A ground rule is derived based on substitutions, where all variables are replaced by constants.
We use uppercase letters for variables and lowercase letters for constants.
Let $B$ be a Herbrand base, including all ground atoms, and an interpretation $I$ is a subset of $B$.  
For an LP $P$ and an interpretation $I$, the immediate consequence operator $T_P:2^B \rightarrow 2^B$ \cite{10.1145/321978.321991} describes the rules with interpretations: 
$T_P(I) = \{head(r) \mid   r \in g(P), body(r) \subseteq I  \}$,
where $g(P)$ is the ground LP based on $P$. 
Hence, given an interpretation $I$ as the current state, $T_P(I)$ is regarded as an interpretation as the next state, which is the set of ground atoms that are derived from the rules of $P$ under the condition that the atoms in $I$ are true.
A forward-chained format \cite{kaminski_eiter_inoue_2018} usually governs the format of an LP, which specifies that the body of rules should satisfy that the variables in the head atom are connected by a binary atomic chain:
\begin{small}
\begin{align}
\label{spat}
    p_t(X,Y) \leftarrow p_1(X, Z_1)\land p_2(Z_1, Z_2)\land \dots \land p_{n+1}(Z_n, Y).
\end{align}
\end{small}
The variable depth represents the number of variables that do not appear in the head atom; e.g., the variable depth is $n$ in the rule (\ref{spat}).

\subsection{Inductive Logic Programming and Propositionalization}

An ILP task aims at generating an LP $P$ headed by a target atom $\alpha_t$ given a tuple $(\mathcal{B}$, $\mathcal{P}$, $\mathcal{N})$. 
Let $p_t$ denote a target predicate; 
$\mathcal{B}$ is a set of ground atoms called background assumptions; $\mathcal{P}$ is a set of positive ground atoms, taken from the ground of the target atom; $\mathcal{N}$ is a set of negative ground atoms, taken outside the ground of the target atom.
Formally, a solution $P$ of an ILP task is:
\begin{align*}
\mathcal{B}, P \models e^{+}, e^+ \in \mathcal{P};  \ \  \mathcal{B}, P \not\models e^{-}, e^- \in \mathcal{N}.
\end{align*}

When learning first-order LPs, the propositionalization method~\cite{Kramer2001} is an effective way to transform relational data into attribute-valued data.
After applying propositionalization methods, we can use NNs to process the attribute-valued data and learn LPs from the relational facts.
In \cite{Franca2014}, the propositionalization method transfers relational facts to unground atoms with Boolean values. 
These unground atoms with values are called first-order features or features for short.

\subsection{Differentiable Semantics of Logic Programs}
\label{PREseman}

In this section, we show the matrix representations and differentiable semantics of LPs.
Let $P$ be an SHLP with a head atom $\alpha_h$, $n$ different body atoms, and $m$ different rules. 
Then $P$ is represented by an SH matrix $\textbf{M}_P \in [0,1]^{m \times n}$.
Each element $a_{kj}$ in $\textbf{M}_P$ is defined as follows~\cite{Gao2021}:
\begin{enumerate}
\item $a_{kj_{i}} = l_i$, where $l_i \in (0,1)$ and $\sum_{s=1}^p l_s  = 1$ 
$(1 \leq i \leq p, \ 1 \leq j_i \leq n, \ 1 \leq  k \leq m)$,  if the $k$-th rule is $\alpha_h \leftarrow \alpha_{j_1} \land \dots \land \alpha_{j_p}$;
\item  $a_{kh} = 1$, if the $k$-th rule is $\alpha_h \leftarrow \alpha_h$;
\item $a_{kj} = 0$, otherwise.
\end{enumerate}
In fact, each row in $\textbf{M}_P$ corresponds to a rule in $P$, and each non-zero value in a row of $\textbf{M}_P$ corresponds to a body atom in the corresponding rule in $P$.
Example~\ref{SHmatrix}
of Appendix~\ref{example_appen}
shows the SH matrix corresponding to an SHLP.
Let $\textbf{M}[k,\cdot]$ and $\textbf{M}'[k,\cdot, \cdot]$ denote the $k$-th row in the matrix $\textbf{M}$ and $k$-th matrix of the three-dimensional tensor $\textbf{M}'$, respectively.
Moreover, an interpretation vector $\textbf{v}=\{a_1, \dots, a_n \}^T$ represents an interpretation in the vector space. If the Boolean value of $\alpha_k$ is \textit{True}, then $a_k = 1$; otherwise, $a_k = 0$.
We also use $\textbf{v}[k]$ to denote the $k$-th element in the vector $\textbf{v}$.
Then, the immediate consequence operator can be represented in the vector space by Equation (\ref{immeope}) \cite{sakama21}.
When $x$ $\geq$ $1$, then the threshold function $\theta(x)$ $=$ $1$; otherwise, $\theta(x)$ $=$ $0$.
To employ an NN to learn the SH matrix of an LP, a differentiable logic semantics used in \cite{Gao2021} replaces the logical or operator with the product t-norm and uses the differentiable function $\phi(x-1)$ in Equation (\ref{act_fun}) to replace the $\theta(x)$ function. The hyperparameter $\gamma$ controls the slope similarity between the functions $\phi$ and $\theta$.
\begin{subequations}
    \noindent\begin{minipage}{0.28\textwidth}
\begin{equation}
    \textbf{v}_o = \vee_{k=1}^m \theta(\textbf{M}_P[k,\cdot] \times \textbf{v}_i^T),  \label{immeope}
\end{equation}
    \end{minipage}%
    \begin{minipage}{0.205\textwidth}
\begin{equation}
    \phi(x) = \frac{1}{1 + e ^{-\gamma x}}. \label{act_fun}
\end{equation}
    \end{minipage}\vskip1em
\end{subequations}

\section{A Neural Network-Based Rule Learner}
\label{Methods}

\begin{algorithm}[tb]
    \caption{The propositionalization method in DFOL}
    \label{palgo}
    \textbf{Input}: A variable set $V=\{X,Y,V_1,V_2,\dots, V_d\}$ with the variable depth $d$; target atom $\alpha_t$, e.g., binary predicate $p_t(X,Y)$ or unary predicate $p_t(X)$; body features set $P_F$; training positive examples $\mathcal{P}$ and training fact set $F$. \\
    \textbf{Output}: A trainable dataset \textit{T}.
    \begin{algorithmic}[1]
    \STATE{\textit{(Preparation Process)}}
    \STATE Let $\mathrm{X}, \mathrm{Y}, \mathrm{V}_1,\mathrm{V}_2, \dots,$ $\mathrm{V}_d$ represent the domains of variables $X, Y, V_1, V_2, \dots, V_d$.
    \STATE If the target predicate $p_t$ is binary, then for each positive example $p_t(o_1,o_2)\in\mathcal{P}$, add $o_1$ and $o_2$ to the sets $\mathrm{X}$ and $\mathrm{Y}$, respectively. Besides, add all entities in $F$ to the sets $\mathrm{V}_1,\mathrm{V}_2,\dots,\mathrm{V}_d $. If $p_t$ is unary, for each positive example $p_t(o_1)\in\mathcal{P}$, add $o_1$ to the set $\mathrm{X}$. Besides, add all entities in $F$ to the sets $\mathrm{Y}$, $\mathrm{V}_1,\mathrm{V}_2,\dots,\mathrm{V}_d$.
    \STATE{\textit{(Generation Process)}} 
    \STATE Calculate the set $S$ with all substitutions: $S = \mathrm{X} \times \mathrm{Y} \times \mathrm{V}_1 \times \mathrm{V}_2 \times  \dots \times \mathrm{V}_d $.
    \FOR{each $\theta_k$$=$$\{x_k/X,y_k/Y,v_1^k/V_1,\dots,v_d^k/V_d\} \in S$}
    \STATE Initialize that $\textbf{v}_i^k$ $=$ $\textbf{0}$ and $\textbf{v}_o^k$ $=$ $[0]$. Under the current substitution $\theta_k$, for each $\alpha_j \in P_F$, $\textbf{v}_i^k[j]=1$ if $g(\alpha_j) \in F$; Then $\textbf{v}_o^k = [1]$ if $g(\alpha_t) \in \mathcal{P}$. Next, add the pair of interpretation vectors $(\textbf{v}_i^k,\textbf{v}_o^k)$ to $T$.
    \ENDFOR
    
    \STATE{\textit{(Examination Process)}}
    \STATE For each data in $T$, delete the data $(\textbf{v}_i^k,\textbf{v}_o^k)$ iff $\textbf{v}_i^k = \textbf{0}$. 
    \STATE Discard the $m$-th body feature and the corresponding values in all input interpretation vectors iff for all $k$ $\in$ $[1, \left\lvert T \right\rvert]$, $ \textbf{v}_i^k[m]$ $=$ $0$ ($m$ $\in$ $[1,\left\lvert P_F\right\rvert]$) holds.
    \STATE \textbf{return} $T$
    \end{algorithmic}
    \end{algorithm}

\subsection{Propositionalization}
In this section, we describe the propositionalization method in Algorithm~\ref{palgo}.
We follow the Closed-World Assumption that any example that does not appear in the set $\mathcal{P}$ is in the set $\mathcal{N}$.
The training fact set $F$ includes training positive examples and training background assumptions.
We assume that each predicate in the task is either binary or unary.
By scanning the dataset once, the number of binary predicates $n_b$ and the number of unary predicates $n_u$ can be determined.
When the body of a rule includes the target atom, the rule is a tautology, which we discard when describing relational facts.
For generating an SHLP $P$ headed by a target atom, the set of possible body features $P_F$ includes all possible binary and unary atoms except the target atom.
Then, we have $\left\lvert P_F\right\rvert  = {\text{A}(\left\lvert V\right\rvert,2) \times n_b + \left\lvert V\right\rvert \times n_u - 1}$, where $\text{A}(\left\lvert V\right\rvert,2)$ is the number of arrangements of $2$ items from $\left\lvert V\right\rvert$ variables.

After the propositionalization, the relational facts are transformed into pairs of interpretation vectors $(\textbf{v}_i, \textbf{v}_o)$.
The features in each $\textbf{v}_i$ are considered as valid features, and let $C$ represent $\left\lvert \textbf{v}_i\right\rvert$.
An input interpretation vector  $\textbf{v}_i$ corresponds to $I_i$, which includes all the values of the valid body features in $P$.
An output interpretation vector $\textbf{v}_o$ corresponds to $I_o$, which determines the value of the head feature in the SHLP $P$. 
Example~\ref{pro_example_pre} 
of Appendix~\ref{example_appen} 
under the \textit{predecessor} (\textit{pre}) relation illustrates the proposed propositionalization method.

We analyze the correctness of the propositionalization method as follows:
In fact, if all the ground body atoms in the LP $P$ describing a relational dataset are satisfied under a substitution $\theta$, the Boolean value of the ground head atom under $\theta$ must be \textit{True}. 
After applying all possible substitutions on all first-order features for a relation dataset,
we can generate pairs of interpretations $(I_i, I_o)$ that correspond to interpretation vectors $(\textbf{v}_i,\textbf{v}_o)$ and satisfy the relation $I_o$ $=$ $T_P(I_i)$.
Therefore, with the help of the robustness of NNs, we can learn the most important body features when the target feature is \textit{True}. 
Moreover, the complexity of the propositionalization method is $O(\left\lvert S \right\rvert \times \left\lvert P_F\right\rvert )$, where $S$ is defined in Algorithm~\ref{palgo}.

\subsection{The Neural Networks in DFOL}
\label{training_pro}
\begin{figure}[tb]
    \centering
    \includegraphics[width=\linewidth]{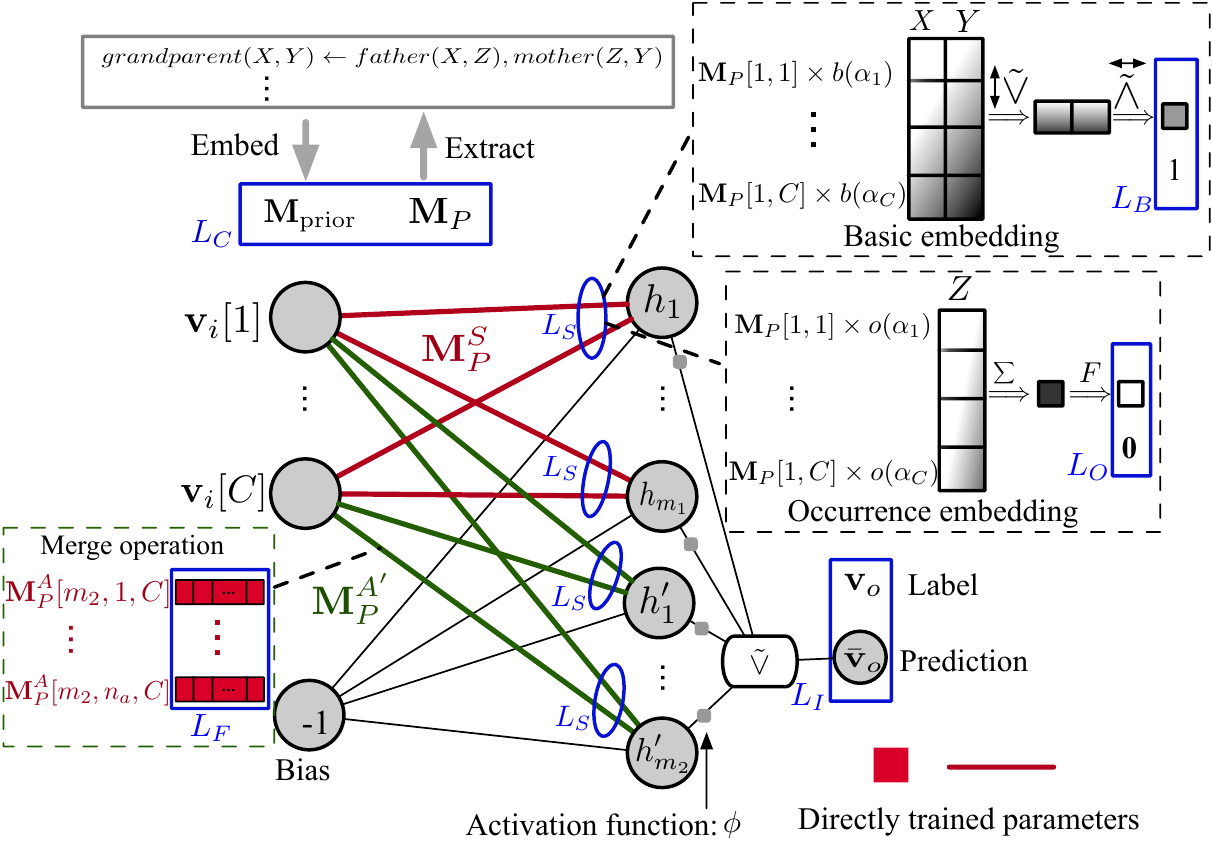}
    \caption{The architecture of DFOL. The logic program, basic and occurence embeddings are instantiated under the \textit{grandparent} task.}
    \label{fig:archi}
\end{figure}

In this section, we describe the proposed NNs and constraint functions depicted in Figure~\ref{fig:archi}. 
Taking the training data $(\textbf{v}_i,$ $\textbf{v}_o)$ $\in$ $T$ as the input, an NN learns the SH matrix $\textbf{M}_P$ encoding an LP $P$.
The LP $P$ meets the forward-chained format described in the rule (\ref{spat}) and the immediate consequence operator $I_o$ $=$ $T_P(I_i)$, where $I_i$ and $I_o$ correspond to $\textbf{v}_i$ and $\textbf{v}_o$, respectively. 
First, we use a matrix $\textbf{M}_P^S \in [0,1]^{m_1\times C}$ as a trainable tensor to encode $P$, where $m_1$ is a hyperparameter describing the number of logic rules.
Besides, we use $\textbf{M}^A_P \in [0,1]^{m_2 \times n_a \times C}$ as another trainable tensor to encode $P$, where $m_2$ and $n_a$ are hyperparameters.
Then, we define the merge operation and concatenation operation  as follows:
\begin{align*}
    \textbf{M}^{A'}_P = \frac{1}{n_a} \sum_{i = 1}^{n_a} \textbf{M}_P^A [\cdot,i,\cdot],  \ 
        \textbf{M}_P = \text{concat}(\textbf{M}_P^{S}, \textbf{M}_P^{A'}),
\end{align*}
where tensors in the arguments of concat function are joined along the vertical dimension. 
Besides, $\textbf{M}_P^{A'}$ $\in$ $[0,1]^{m_2 \times C}$ and $\textbf{M}_P$ $\in$ $[0,1]^{m\times C}$, where $m = m_1+m_2$.
We illustrate the merge and concat functions in Example~\ref{mere_examle} 
of Appendix~\ref{example_appen}.
We replace the immediate consequence operator of $P$ with a differentiable inference and define an inference loss as follows:
\begin{align*}
\bar{\textbf{v}}_o =   \widetilde{\vee}_{k=1}^m (\phi(\textbf{M}_P[k,\cdot] \times \textbf{v}_i^{T} - 1)), \ 
L_I = H(\bar{\textbf{v}}_o, \textbf{v}_o), 
\end{align*}
where the activation function $\phi$ is defined in Equation (\ref{act_fun}).
The function $H$ denotes the binary cross-entropy function. 
The symbol $\Tilde{\lor}$ denotes the differentiable fuzzy or operation \cite{DBLP:books/kl/Hajek98}.
Compared with other fuzzy logic semantics, we use product t-norm as the fuzzy logic semantics in DFOL for avoiding a zero gradient~\cite{evans2018learning}:
\begin{align*}
    \widetilde{\vee}_{i=1}^{n} x_i = 1-  \prod_{i=1}^n  (1-x_i), \ \ 
    \widetilde{\wedge}_{i=1}^{n} x_i = \prod_{i=1}^n x_i
\end{align*}

Now, we analyze the roles of the tensor $\textbf{M}_P^A$. 
Compared with $\textbf{M}_P^S$, $\textbf{M}_P^A$ has more parameters. 
Besides, the merge operation keeps the numbers of columns including the non-zero values from $\textbf{M}_P^A[k,\cdot,\cdot]$ to $\textbf{M}_P^{A'}[k,\cdot]$.
Thus, when $\textbf{M}_P^{A'}[k,\cdot]$ encodes a correct rule $r_k$ headed by the target atom, the rows in the matrix $\textbf{M}_P^A[k,\cdot,\cdot]$ have the opportunity to represent rules headed by auxiliary predicates~\cite{Muggleton2015}.
The body of a rule headed by an auxiliary predicate may include a part of the body atoms that appear in the rule $r_k$.
Hence, the matrix $\textbf{M}_P^A$ boosts the training process. 
Example~\ref{auxi_inven}
of Appendix~\ref{example_appen}
describes the process of auxiliary predicate invention. 
Hence, the parameters $m_2$ and $n_a$ indicate the number of rules headed by the target atom and the number of rules headed by auxiliary predicates, respectively.

Then, we describe other essential constraint functions to generate precise LPs.
Firstly, we use a sum loss function $L_S$ according to the property described in Section~\ref{PREseman} that the sum of each row in an SH matrix is equal to one:
\begin{align*}
    L_S = \sum_{k=1}^{m}  \text{MSE} (\sum_i^{C} \textbf{M}_P[k ,i], {1}),
\end{align*}
where $\text{MSE}$ is the mean square error loss function. 
To make the rules extracted from $\textbf{M}_P$ meet the forward-chained format defined in the rule (\ref{spat}), we stipulate two spatial constraints:
\begin{enumerate}
    \item Basic constraint: The body of each rule in an LP contains all variables that appear in the head atom. 
    \item Occurrence constraint:     
    In the body of each rule, the number of occurrences of each variable that does not appear in the head atom is not one.
\end{enumerate}
Then, we give each body atom a basic embedding and an occurrence embedding, and devise basic loss function and occurrence loss function to implement the above spatial constraints.
Suppose that the arity of the head atom in an LP $P$ is $t$, then the variable depth is $\left\lvert V\right\rvert-t$.
Let $V_h$ and $V_o$ be the variable sets with elements appearing in the head atom and not appearing in the head atom, respectively. 
Let $b(\alpha)$ $\in$ $\{0,1\}^t$ and $o(\alpha)$ $\in$ $\{0,1\}^{\left\lvert V\right\rvert- t}$ be the basic and occurrence embeddings corresponding to the body atom $\alpha$, respectively. 
If the $i$-th variable in $V_h$ (or $V_o$) appears in $\alpha$, then $b(\alpha)[i]$$=$$1$ (or $o(\alpha)[i]$$=$$1$); otherwise,  $b(\alpha)[i]$$=$$0$ (or $o(\alpha)[i]$$=$$0$). 
Example~\ref{bandoemb} 
of Appendix~\ref{example_appen} 
illustrates a basic embedding and an occurrence embedding.
Then, we devise the following functions to implement the basic constraint:
\begin{align*}
\textbf{M}_{b}^k = \text{concat}( &\textbf{M}_P[k,1]  \times b(\alpha_1), \textbf{M}_P[k,2]  \times b(\alpha_2), \dots, \nonumber
\\ & \textbf{M}_P[k,C]  \times b(\alpha_C)) 
\end{align*}
where $\alpha_i$ is the $i$-th valid first-order feature. 
The matrix $\textbf{M}_b^k$ $\in$ $[0,1]^{C\times t}$ includes the occurrence information of all variables in $V_h$ across all valid features for the $k$-th rule in an LP. 
Then, we use the fuzzy conjunction and disjunction operators to calculate the possibility that all variables in $V_h$ appear in the $k$-th rule at once, and we define a basic loss function $L_B$:
\begin{equation*}
    \text{basic}_k =  \widetilde{\wedge}_{j=1}^{t }  \widetilde{\vee}_{i=1}^{C}  \textbf{M}_b^k[i,j], \ 
    L_{B} = \sum_{k=1}^{m} \text{MSE}(\text{basic}_k, 1).
\end{equation*}
To implement the occurrence constraint, we use the following equation to concatenate all possibilistic occurrence embeddings across all valid features to $\textbf{M}_{o}^k$ $\in$ $[0,1]^{C \times (\left\lvert V\right\rvert - t)}$:
\begin{align*}
    \textbf{M}_{o}^k = \text{concat}(&\textbf{M}_P[k,1]\times o(\alpha_1),\textbf{M}_P[k,2]\times o(\alpha_2),  \dots,
    \\ & \textbf{M}_P[k,C]\times o(\alpha_C) ).
\end{align*}
Then, the possibility for each variable in $V_o$ across all the valid features in the $k$-th rule are summarized into the matrix $ \textbf{V}_{o}^k$ $=$ $\sum_{i=1}^{C} \textbf{M}_{o}^k[i,\cdot]$.
Hence, $\textbf{V}_{o}^k \in [0,1]^{\left\lvert V\right\rvert -t}$ includes the possibility of each variable in $V_o$ in the $k$-th rule.
Then, we set a measurement function $F(x): \mathbb{N}  \to \left[ 0,a \right]$ to reflect the number of occurrences of each variable in $V_o$, and define an occurrence loss function $L_O$ as follows:
\begin{align*}
     \ F(x) = a\cdot e^{b-c (x-d)^2}, \ L_{O} = &\sum_{k=1}^{m}  F(\textbf{V}_{o}^k),
\end{align*}
where $a$, $b$, $c$, and $d$ are hyperparameters.
As $x$ gets closer to $d$, the value of $F(x)$ gets larger; otherwise, $F(x)$ gets closer to $0$.
Example~\ref{operations_in_formula} 
of Appendix~\ref{example_appen} 
illustrates the operations related to a basic and occurrence embedding.

Next, we consider the cosine similarity, $\cos (\textbf{u},\textbf{v}) = \frac{\textbf{u} \cdot \textbf{v} }{ \mid \textbf{u}\mid \mid\textbf{v}\mid}$, denoting the similarity between two vectors. 
When two vectors have greater dissimilarity, their cosine similarity is close to -1; otherwise, their cosine similarity is close to 1. 
Since the parameter $n_a$ in the matrix $\textbf{M}_P^{A}$ indicates the number of rules headed by auxiliary predicates, we reduce the cosine similarity of each 2-combinations of all rows in the matrix $\textbf{M}_P^A[k,\cdot,\cdot]$ for generating more possible formats of rules headed by the auxiliary predicates.
Then, we calculate the loss function $L_F$:
\begin{equation*}
\resizebox{\linewidth}{!}{$
    \displaystyle
    L_F  = \sum_{k=1}^{m_2}  \sum_{{(i_1,i_2) \in \binom{[1,n_a]}{2} }} \text{MSE}( \cos (\textbf{M}_P^A[k, i_1 ,\cdot],  \textbf{M}_P^A[k,i_2, \cdot]), -1), 
$}
\end{equation*}
where $\binom{[1,n_a]}{2}$ is the set of all 2-combinations of the integer set $[1,n_a]$.
To apply the curriculum learning: The system consolidates what it learns in one episode, storing it as background knowledge, and reusing it in subsequent episodes~\cite{DBLP:journals/ai/EvansHWKS21}.
We devise a strategy to implement the curriculum learning that DFOL uses sound logic rules extracted in every few epochs as the prior knowledge to reduce the search space in the following epochs. 
We reduce the cosine similarity between the rows in the trainable matrix $\textbf{M}_P$ and a learned matrix  $\textbf{M}_{\text{prior}}$ $\in$ $[0,1]^{m_p \times C}$ corresponding to the sound rules described in Section~\ref{rule_extraction}, where $m_p$ is the number of extracted sound rules. 
The loss function $L_C$ is defined as follows:
\begin{align*}
    L_C   =  \sum_{\mathclap{(k_1, k_2) \in [1,m] \times  [1,m_p]}} \text{MSE}(  \cos (\textbf{M}_P[k_1,\cdot],   \textbf{M}_{\text{prior}}[k_2, \cdot]), -1). 
\end{align*}

In summary, the final loss is the weighted sum of the losses in $ \textbf{L}$$=$$[L_I, L_S, L_B, L_O, L_F, L_C]$, i.e., 
$\text{loss}$ $=$ $\varTheta$ $\cdot$ $\textbf{L}$,
where $\varTheta$ is a hyperparameter vector.
We use the Adam algorithm~\cite{DBLP:journals/corr/KingmaB14} to minimize the final loss.

\subsection{Rule Extraction}
\label{rule_extraction}
In this section, we describe how to extract LPs from a trained SH matrix and the definition of the precision of a rule. 

In a trained SH matrix $\textbf{M}_P$, the element at the $m$-th row and $n$-th column represents the possibility of $n$-th valid body feature in the $m$-th rule $r_m$ in $P$.
We use multiple thresholds called rule filters, denoted as $\tau_f$, on $\textbf{M}_P$ to extract rules.
Let rule filters range from 0 to 1 with step 0.05, and let $\mathsf{T}$ be the set with all rule filters.
For a $\tau_f$, we let the valid features in the $k$-th row of $\textbf{M}_P$ with values greater than $\tau_f$ be the elements in the $body(r_k)$.
We iteratively apply each $\tau_f$ on the trained matrix $\textbf{M}_P$, and a rule set $\tilde{R}$ with $m$$\times$$\left\lvert \mathsf{T} \right\rvert$ rules is generated.
Next, we describe the definition of the precision of a rule.  
Let $n_{r}$ be the number of the substitutions that satisfy both the body and the head atom of a rule, and let $n_{b}$ be the number of the substitutions that satisfy only the body of a rule, where the substitutions are computed based on the Datalog and seen facts.
Then, we regard the ratio $\frac{n_{r}}{n_{b}}$ as the precision of the rule $r$. 
A rule is correct with the precision value 1, and a rule is incorrect with the precision value 0. 
If a precision value floats within the interval $(0,1)$, then the rule may also be correct due to the incompleteness of the seen facts. 
Therefore, we set another threshold called soundness filter $\tau_s$. 
Rules with precision values no lower than $\tau_s$ are called sound rules and are added to $P$ from the set $\tilde{R}$.
To use the curriculum learning strategy, we concentrate the trained rows in $\textbf{M}_P$ corresponding to the sound rules into $\textbf{M}_{\text{prior}}$ after every few epochs. 
The matrix $\textbf{M}_{\text{prior}}$ is considered as the prior knowledge to boost the training process in the following training epochs. 
When all training epochs are finished, the sound LP is stored in $P$.

\section{Experimental Evaluation}
\label{Experimental}
In this section, we present the performance of DFOL and make comparisons with NTP$\lambda$ \cite{DBLP:conf/nips/Rocktaschel017}, $\partial$ILP \cite{evans2018learning}, and NeuralLP \cite{YangFanetc}.
The ratio of positive test examples covered by an LP $P$ to all positive test examples is used as the accuracy of $P$, which is also the recall value of $P$ in the test dataset.
Due to the small size of trainable matrices in DFOL and to demonstrate the efficiency of DFOL, all experiments are executed on 24GB of memory and an 8-core Intel i7-6700 CPU. 
We limited the running time to one hour when generating an SHLP. Besides, the variable depth in each task does not exceed two.

\begin{figure}[tb]
    \begin{minipage}{.12\textwidth}
        \centering
        \includegraphics[width=\linewidth]{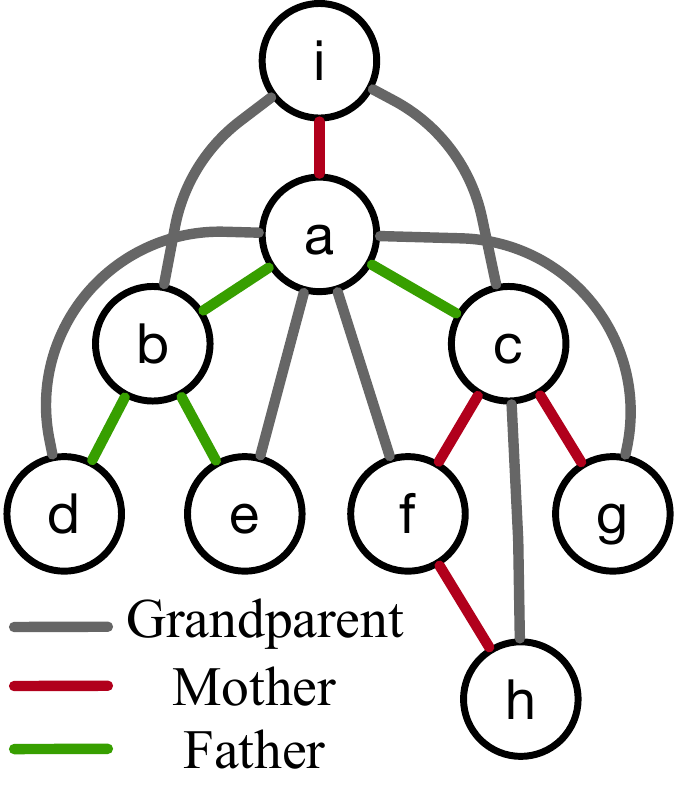}
        \caption{The background assumptions and the positive examples in the \textit{grandparent} task.}
        \label{fig:bk}
    \end{minipage}   \   { } \  \ \ 
    \begin{minipage}{.33\textwidth}
        \includegraphics[width=\linewidth,right]{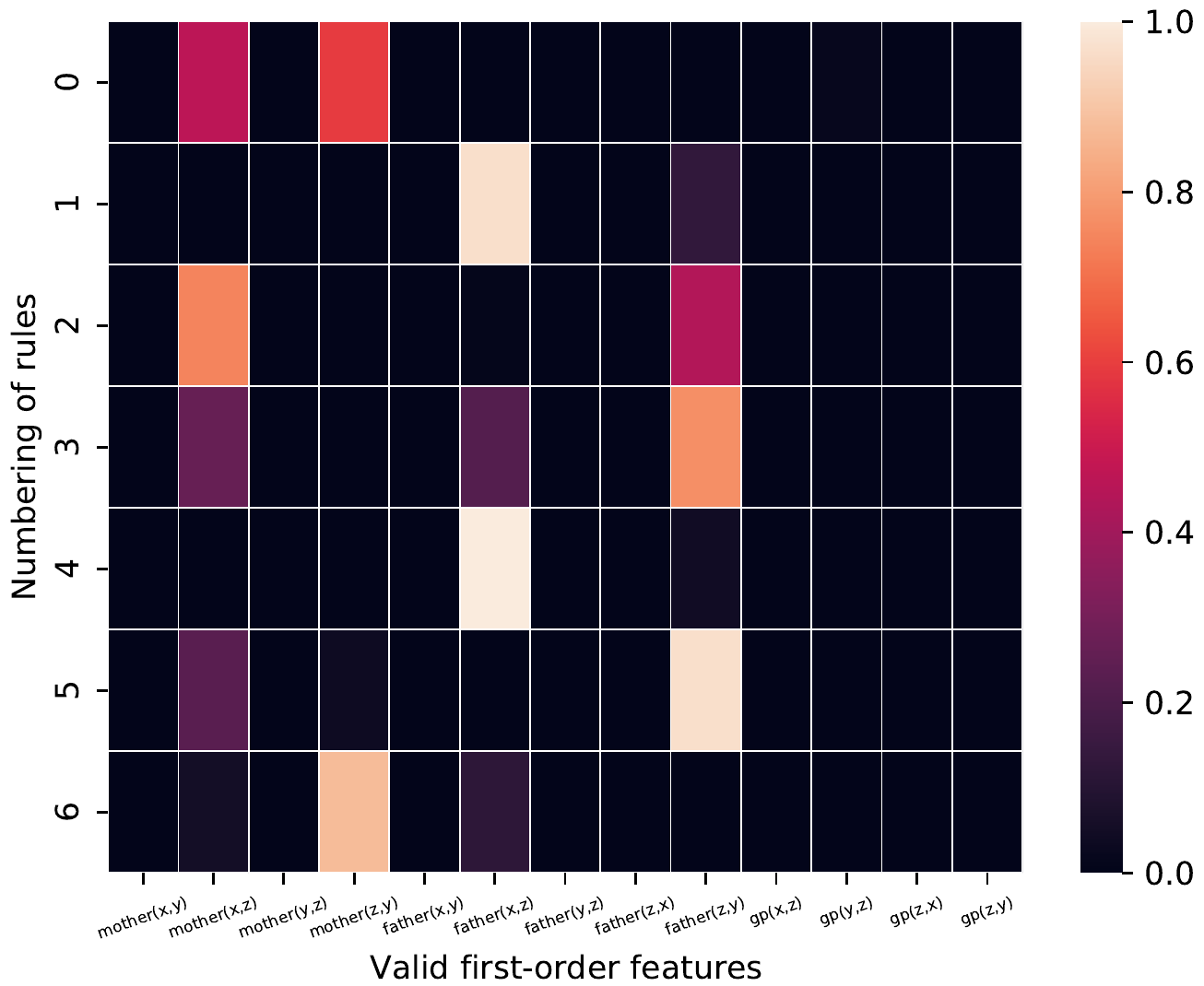}
        \caption{The learned matrix in the \textit{grandparent} task.}
        \label{fig:matrix}
    \end{minipage}
\end{figure}

\subsection{Learning from ILP Datasets}
\label{learnILP}

In this section, we present the results of DFOL on 20 classifications of ILP datasets~\cite{evans2018learning}.
Most of the ILP datasets have small numbers of data.
In these benchmarks, we set $\tau_s = 1$. 
Hence, the precision value of each generated rule is 1. 
The accuracy values of all the generated LPs are 100\% in this section. 
The task definitions and generated solutions are shown in Appendix~\ref{logic_program_app}. 
We use the \textit{grandparent} (\textit{g}) relation as an example to show the results.
The background assumptions include the facts with \textit{mother} (\textit{m}) and \textit{father} (\textit{f}) relations, and the positive examples include the facts with \textit{g} relation.
The training facts are presented in Figure~\ref{fig:bk}. 
The learned matrix $\textbf{M}_P$ is presented in Figure~\ref{fig:matrix} and the extracted LP $P$ with 100\% accuracy is shown as follows:
\begin{small}
\begin{align*}
    g(X,Y) \leftarrow m(X,Z) \land m(Z,Y), \ 
    g(X,Y) \leftarrow f(X,Z) \land f(Z,Y) \\
    g(X,Y) \leftarrow m(X,Z) \land f(Z,Y),\ 
    g(X,Y) \leftarrow f(X,Z) \land m(Z,Y)
\end{align*}
\end{small}

Based on the comparison with $\partial$ILP and NeuralLP, DFOL completes 20 of the 22 benchmarks and outperforms NeuralLP in 19 of all benchmarks.
Since $\partial$ILP is a memory expensive model~\cite{evans2018learning}, $\partial$ILP cannot generate correct LPs in the case of a large number of objects, e.g., in the even number (20), graph coloring (10) tasks, and knowledge base dataset considered in the paper.
Besides, $\partial$ILP requires a logical template, a more specific prior knowledge than the forward chain format.
Hence, DFOL is a precise rule learner for smaller ILP datasets.   
However, DFOL runs over the memory limit when producing all substitutions in the propositionalization on the \textit{husband} and \textit{uncle} tasks.

\subsection{Learning from Ambiguous Datasets}
\label{ambi}
In this section, we test the robustness of DFOL and consider two cases: (1) learning from probabilistic facts; (2) learning from mislabeled facts.
We set the soundness filter $\tau_s$ to 1 to obtain rules with precision values of 1.

For the first case, each ground atom in the training fact set $F$ and training negative example set $\mathcal{N}$ has a probability.
Let $\epsilon \sim  N(0,\sigma^2)$, then the probability of fact in $F$ satisfies the distribution $p^+_i$ $=$ $min(1 - \epsilon, 1)$, and the probability of ground atom in $\mathcal{N}$ satisfies the distribution $p^-_i = max(\epsilon, 0)$. 
The standard deviation $\sigma$ ranges from 0.5 to 3 with step 0.5. 
We present the distribution of the positive and negative examples on \textit{lessthan} (\textit{lt}) and \textit{pre} tasks when $\sigma = 3$ in Figures~\ref{fig:dis_less} 
and \ref{fig:dis_pre} 
of Appendix~\ref{figdis}.
From the two distributions, we derive that the values of examples are sufficiently ambiguous when $\sigma=3$.
For the second case, both positive and negative training examples are mislabeled with the mutation rate $\mu$ ranging from 0.05 to 1 with step 0.05.
In Table~\ref{ambigoues}, we show the maximum standard deviation $\sigma$ and mutation rate $\mu$ when the accuracy values of the generated LPs are 100\%.
From the results, we conclude that DFOL is robust when handling ambiguous data.
When $\mu = 1$, the semantics of head predicate in \textit{lt} task is \textit{largethan} relation, and the result is:
\begin{small}
    \begin{align*}
        lt(X,Y) \leftarrow succ(Y,X), \ 
        lt(X,Y) \leftarrow lt(X,Z) \land lt(Z,Y).
    \end{align*}
\end{small}

\begin{table}[tb]
    \caption[]{The results on ambiguous datasets. The notations \textit{Con} and \textit{DE} represent \textit{connectedness} and \textit{directed edge} tasks, respectively. }
    \label{ambigoues}
    \centering
    \begin{tabular}{lrrrrrr}
    \toprule
    & \textit{Lt} & \textit{Pre} & \textit{Member} & \textit{Son}  & \textit{Con} & \textit{DE} \\
    \midrule
    $\sigma$ & 3           & 3            & 3               & 3    & 2           & 3                        \\
    \midrule
    $\mu$     & 0.95        & 0.95         & 0.90              & 0.95 & 0.95          & 0.95                    \\
    \bottomrule
    \end{tabular}
\end{table}

\subsection{Learning from Knowledge Bases}
\label{learnknowge}

\begin{table}[t]
    \caption[]{Comparison on knowledge bases. The results in bold indicate the highest accuracy on the corresponding test datasets. The ACC@S$n$ represent the accuracy of the generated LP on the S$n$ subset of Countries. 
    The ACC@\textit{h} represent the accuracy of LP with the head predicate $h$. Besides, \textit{blo}, \textit{int}, \textit{neg}, and \textit{intw} denote the relations of \textit{blockpositionindex}, \textit{intergovorgs3}, \textit{negativecomm}, and \textit{interacts\_with}.}
    \centering
    \label{large_base_compare}
    \begin{tabular}{llrrr}
    \toprule
    {Dataset}   & Metrics   & NTP$\lambda $         & NeuralLP & DFOL           \\ 
    \midrule
    {Countries} & {ACC}@S1               & \textbf{100.00} & \textbf{100.00}    & \textbf{100.00}   \\
                &  ACC@S2                    & \textbf{100.00} & \textbf{100.00}    & \textbf{100.00}   \\
                &  ACC@S3                     & \textbf{100.00} & $- $     & $- $          \\ 
    \midrule
    {Nations}   & MRR & 41.79 &  56.49   &  \textbf{78.88}  \\
     & HITS@1 & 41.79 &   52.49  & \textbf{73.88}   \\
     & HITS@3 & 41.79 &  60.95   &  \textbf{84.58}  \\
     & HITS@10 & 41.79 &  61.19   &  \textbf{85.07}  \\
    &  ACC@\textit{blo} & \textbf{100.00} & 50.00    & \textbf{100.00}   \\
    &   ACC@\textit{int}    & \textbf{84.62}        & \textbf{84.62}    & \textbf{84.62} \\
    &   ACC@\textit{neg}    & 37.50         & \textbf{75.00}    & \textbf{75.00} \\ 
    \midrule
    {UMLS}      
    & MRR & 30.03 &  66.69   &   \textbf{74.96} \\
    & HITS@1 & 29.95  & 61.27  &  \textbf{71.41}  \\
    & HITS@3 & 30.11 &72.31  & \textbf{78.82}   \\
    & HITS@10 & 30.11 &72.31  &  \textbf{78.97}  \\
    &   ACC@\textit{isa}      & 65.96        & 63.83     & \textbf{91.48} \\
    & ACC@\textit{intw}  & 83.67        & 86.67    & \textbf{100.00} \\
    \bottomrule
    \end{tabular}
\end{table}

In this section, we test DFOL on three real knowledge bases with large data sizes, including Countries dataset \cite{DBLP:conf/aaaiss/Bouchard0T15}, Unified Medical Language System (UMLS), and Nations dataset \cite{DBLP:conf/icml/KokD07}. 
We set $\tau_s = 0.3$ because the training facts are incomplete in the realistic scenario. 
Hence, we present each rule with the value of precision, $n_{r}$ and $n_b$, i.e., $(\frac{n_{r}}{n_b},n_{r},n_b)$ in Appendix~\ref{logic_program_app}.
The descriptions of these datasets are presented in Table~\ref{datasetd} 
of Appendix~\ref{figdis}.
For the Countries dataset, training facts are split into S1, S2, and S3 sub-datasets. 
From the sub-datasets S1 to S3, the learning difficulty is increasing because the related relations are missing corresponding to the test cities \cite{DBLP:conf/nips/Rocktaschel017}.
For the Nations and UMLS datasets, we divide each dataset into 80\% training facts, 10\% development facts, and 10\% test facts.
We compare LPs generated by DFOL, NTP$\lambda$, and NeuralLP, through the indicators of accuracy, mean reciprocal rank (MRR), and HITS@m \cite{DBLP:conf/nips/BordesUGWY13} in Table~\ref{large_base_compare}. 


From the results in Table~\ref{large_base_compare}, we conclude that DFOL has better performance than other baselines in general.
Thus, we show that DFOL guarantees the accuracy of the generated LPs, as well as the scalability and interpretability on larger relational datasets.
In addition, in all sub-datasets of the Countries dataset, although facts with \textit{locatedIn} and \textit{neighbor} predicates related to the test countries are missing, these facts are still kept in the training facts that are not related to the test countries. 
Hence, we can extract the same LP from all tasks.
The generated symbolic LP can describe all the test facts in both S1 and S2 sub-datasets.
However, because of the missing related facts in the S3, the 100\% accuracy logic rule needs at least four variables.
DFOL runs out of memory when generating all substitutions for four variables in the propositionalization process on the S3 sub-dataset.

\section{Conclusion and Future Work}
\label{Conclusion}
In this paper, we proposed differentiable first-order rule learner (DFOL), which learns first-order logic programs in the forward-chained format from relational facts without logic templates. 
Through the proposed propositionalization method, DFOL translates relational data to the neural network-readable data. 
By applying the proposed constraint functions, DFOL can learn correct first-order logic programs with a few trainable parameters.
We extract symbolic logic programs from the trainable matrices directly and apply prior knowledge as constraints to implement curriculum learning.
Experimental results indicate that DFOL can learn first-order logic programs with both high accuracy and precision from several standard inductive logic programming tasks, probabilistic relational facts, and knowledge bases. 
Hence, DFOL can learn rules from relational facts in a flexible, precise, robust, scalable, and computationally cheap manner.

In the future, we will try a more efficient propositionalization method in order to learn relational tasks with larger entities. 
Besides, we will extend DFOL to an explainable deep neural network framework to generate logic programs with more variables. We also plan to support the function in the term and the negation in the body of a rule.

\bibliographystyle{named}
\bibliography{ijcai22}

\appendix
\section{Examples}\label{example_appen}

The examples mentioned in the paper are presented as follows:

\begin{example}[Same head matrix]
    \label{SHmatrix}
	Consider an SH program $P$ with three rules and $body(P)$ $=$ $\{$$p(X,Z)$$,$ $p(Y,X)$$,$ $p(Y,Z)$$,$ $p(Z,X)$$,$ $p(Z,Y)$$\}$, and the logic program $P$ is:
	\begin{align*}
			p(X,Y)  &  \leftarrow p(Y,Z) \land p(Z,X). \\
			p(X,Y)  &  \leftarrow p(X,Z) \land p(Z,Y). \\
			p(X,Y)  &  \leftarrow p(Y,X).  
    \end{align*}
    Then, one of SH matrix $\textbf{M}_P$ of the logic program $P$ is presented as follows:
    \begin{align*}
		\mathbf{M}_P = 
		\begin{bmatrix}
		0 & 0 & \frac{1}{2} & \frac{1}{2} & 0 \\
		\frac{1}{2} & 0 & 0 & 0 & \frac{1}{2} \\
		0 & 1 & 0 & 0 & 0
		\end{bmatrix}.
	\end{align*}
\end{example}

\begin{example}[The result after the propositionalization method]
    \label{pro_example_pre}
    In the \textit{predecessor} example, assume that entity set is $E$ $=$ $\{$$0$$,$ $1$$,$ $2$$\}$. The background assumptions $\mathcal{B}$ and the positive examples $\mathcal{P}$ are $\{$$succ(0,1)$$,$ $succ(1,2)$$\}$ and $\{$$pre(1,0)$$,$ $pre(2,1)$$\}$, respectively.
    Assume that the variable set is $V$ $=$ $\{$$X$$,$ $Y$$\}$, then the possible body feature set $P_F$$=$$\{$$succ(X,Y)$$,$ $succ(Y,X)$$,$ $pre(Y,X)$$\}$. 
    According to the positive examples, the domians $\mathrm{X}$$=$$\{1,2\}$ and $\mathrm{Y}$$=$$\{0,1\}$.
    Then, the substitution set is:$\{\{1/X,0/Y\}$, $\{1/X,1/Y\}$, $\{2/X,0/Y\}$, $\{2/X,1/Y\}\}$.
    Bofore the examination process, the trainable dataset $T$ is:$\{([0,1,0],[1]),([0,0,0],[0]), ([0,0,0], [0]), ([0,1,0],[1])\}$
    After the examination process, only the valid body predicate $succ(Y,X)$  is kept, and $T$ $ = \{([1],[1])\}$. 
\end{example}

\begin{example}[Merge operation and concat functions]
    \label{mere_examle}
    Assume that the target atom is $p$$(X$$,$ $Y)$. The variable set is $\{$$X$$,$ $Y$$,$ $Z$$\}$, and the valid body feature set $P_v$ is: $\{$$p(X,Z)$$,$ $p(Y,X)$$,$ $p(Y,Z)$$,$ $p(Z,X)$$,$ $p(Z,Y)$$\}$. Hence, the $V_h = \{X,Y\}$ and $V_o$ $=$$\{Z\}$. 
    Assume that the matrices $\textbf{M}_P^A \in [0,1]^{1 \times 2 \times 5 }$ and $\textbf{M}_P^S \in [0,1]^{1\times 5}$ are:
    \begin{align*}
        \mathbf{M}_P^A[1,\cdot,\cdot] &= 
		\begin{bmatrix}
		0.01 & 0.90 & 0.00 & 0.04 & 0  \\
		0.05 & 0.80 & 0.20 & 0.00 & 0   
		\end{bmatrix}, \\ 
        \mathbf{M}_P^S &= 
		\begin{bmatrix}
		0 & 0.95 & 0 & 0.03 & 0.02 
		\end{bmatrix} .
    \end{align*}
    Then, the matrices $\textbf{M}_P^{A'}\in[0,1]^{1\times 5}$ and $\textbf{M}_P \in [0,1]^{2\times 5}$ are presented as follows:
    \begin{align*}
        \mathbf{M}_P^{A'} &= 
		\begin{bmatrix}
		0.03 & 0.85 & 0.10 & 0.02 & 0 
		\end{bmatrix}, \\ 
        \mathbf{M}_P &= 
		\begin{bmatrix}
        0 & 0.95 & 0 & 0.03 & 0.02 \\
        0.03 & 0.85 & 0.10 & 0.02 & 0  
		\end{bmatrix} .
    \end{align*}

\end{example}

\begin{example}[Auxiliary predicates invention]
\label{auxi_inven}
We consider the following logic program $P$ describing the target predicate \textit{grandparent} with the body predicates \textit{mother} and \textit{father}:
\begin{align*}
    grandparent(X,Y) &\leftarrow mother(X,Z), father(Z,Y).
\end{align*}
Assume the valid body feature set is equal with all body feature set $P_F$ $=$ $\{$$mother(X,Y)$$,$ $mother(X,Z)$$,$ $\dots$ $,$ $father(Z,Y)$$,$ $\dots$ $,$ $grandparent(Z,Y)$$\}$.
Then, one of SH matrix $\textbf{M}_P\in[0,1]^{1\times 17}$ corresponding the program $P$ is:
\begin{equation*}
    (\mathbf{M}_P)^T = 
    \begin{matrix}
        mother(X,Y) & 0 \\
        mother(X,Z) & 0.5 \\
        \vdots  & \vdots \\
        father(Z,Y) & 0.5\\
        \vdots & \vdots \\
        grandparent(Z,Y) & 0
    \end{matrix}.
\end{equation*}
If we do not consider the matrix $\textbf{M}_P^S$, then a possible $\textbf{M}_P^A\in[0,1] ^{ 1\times 2 \times 17}$ is:
\begin{equation*}
    (\mathbf{M}_P^A[1,\cdot, \cdot])^T = 
    \begin{matrix}
        mother(X,Y) & 0 & 0 \\
        mother(X,Z) & 1 & 0\\
        \vdots  & \vdots & \vdots\\
        father(Z,Y) & 0 & 1\\
        \vdots & \vdots & \vdots\\
        grandparent(Z,Y) & 0 & 0
    \end{matrix}.
\end{equation*}
The logic programs encoded in the metrix $\textbf{M}_P^A$ are interpreted as:
\begin{align}
    parent(X,Z) &\leftarrow mother(X,Z), \label{aux_head_1} \\ 
    parent(Z,Y) &\leftarrow father(Z,Y). \label{aux_head_2}
\end{align}
The rules~(\ref{aux_head_1}) and~(\ref{aux_head_2}) are regarded as rules headed by an auxiliary predicate \textit{parent}.

\end{example}

\begin{example}[Basic embeddings and occurrence embeddings]
\label{bandoemb}

We use the target atom, variable set, and valid body feature set in Example~\ref{mere_examle}, the basic embeddings of all the body features are: $b(P_v)$ $=$ $\{$$[1,0]$$,$ $[1,1]$$,$ $[0,1]$$,$ $[1,0]$$,$ $[0,1]$$\}$, and the occurrence embeddings for all the body features are: $o(P_v)$$ = \{[1], [0],[1],[1],[1]\}$.
\end{example}

\begin{example}[The operations related to basic and occurrence embeddings]
    \label{operations_in_formula}
    We use the target atom, variable set, and valid body feature set in Example~\ref{mere_examle}. 
    Then, the matrices $\textbf{M}_b^1$ and $\textbf{M}_b^2$ are: 
\begin{equation*}
    \mathbf{M}_b^1 = 
    \begin{bmatrix}
    0&0\\
    0.95&0.95\\
    0&0\\
    0.03&0\\
    0&0.02 
    \end{bmatrix}, \ 
    \mathbf{M}_b^2 = 
    \begin{bmatrix}
        0.03&0\\
        0.85&0.85\\
        0&0.10\\
        0.02&0\\
        0&0 
    \end{bmatrix}.
\end{equation*}
Hence, we get $\text{basic}_1 = (1-0.05\times 0.97)\times (1- 0.05 \times 0.98) = 0.90$ and $\text{basic}_2 = (1-0.97\times 0.15\times 0.98)\times (1- 0.15 \times 0.90) = 0.74$.  

In addition,  the matrices $\textbf{M}_o^1$ and $\textbf{M}_o^2$ are: 
\begin{align*}
    \mathbf{M}_o^1 &= 
    \begin{bmatrix}
0 & 0 & 0 & 0.03 & 0.02
    \end{bmatrix}^T, 
    \\
    \mathbf{M}_o^2 &= 
    \begin{bmatrix}
0.03 & 0& 0.10& 0.02& 0
    \end{bmatrix}^T.
\end{align*}
When the hyperparameters $a = b = d = 1$, and $c=10$, then $L_O = [e^{1-10\times (0.05-1)^2}] + [ e^{1-10\times (0.15-1)^2} ] =  [2.31 \times 10^{-3}]$.

\end{example}

\section{Illustrations}\label{figdis}

\begin{figure}[H]
        \centering
        \includegraphics[width=\linewidth]{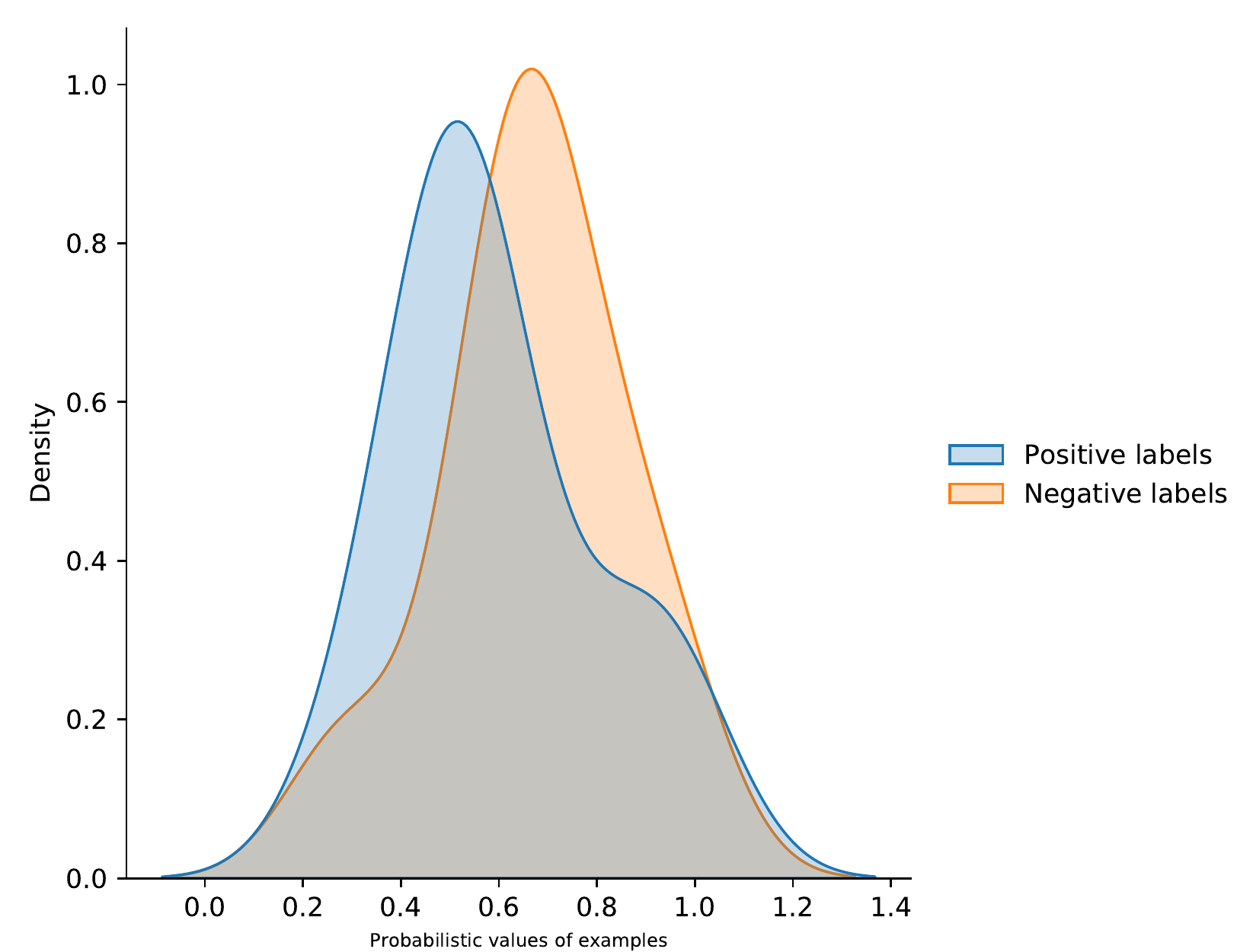}
        \caption{The distribution of examples in the \textit{lessthan} task when $\sigma = 3$.}
        \label{fig:dis_less}
\end{figure}

\begin{figure}[H]
        \centering
        \includegraphics[width=\linewidth]{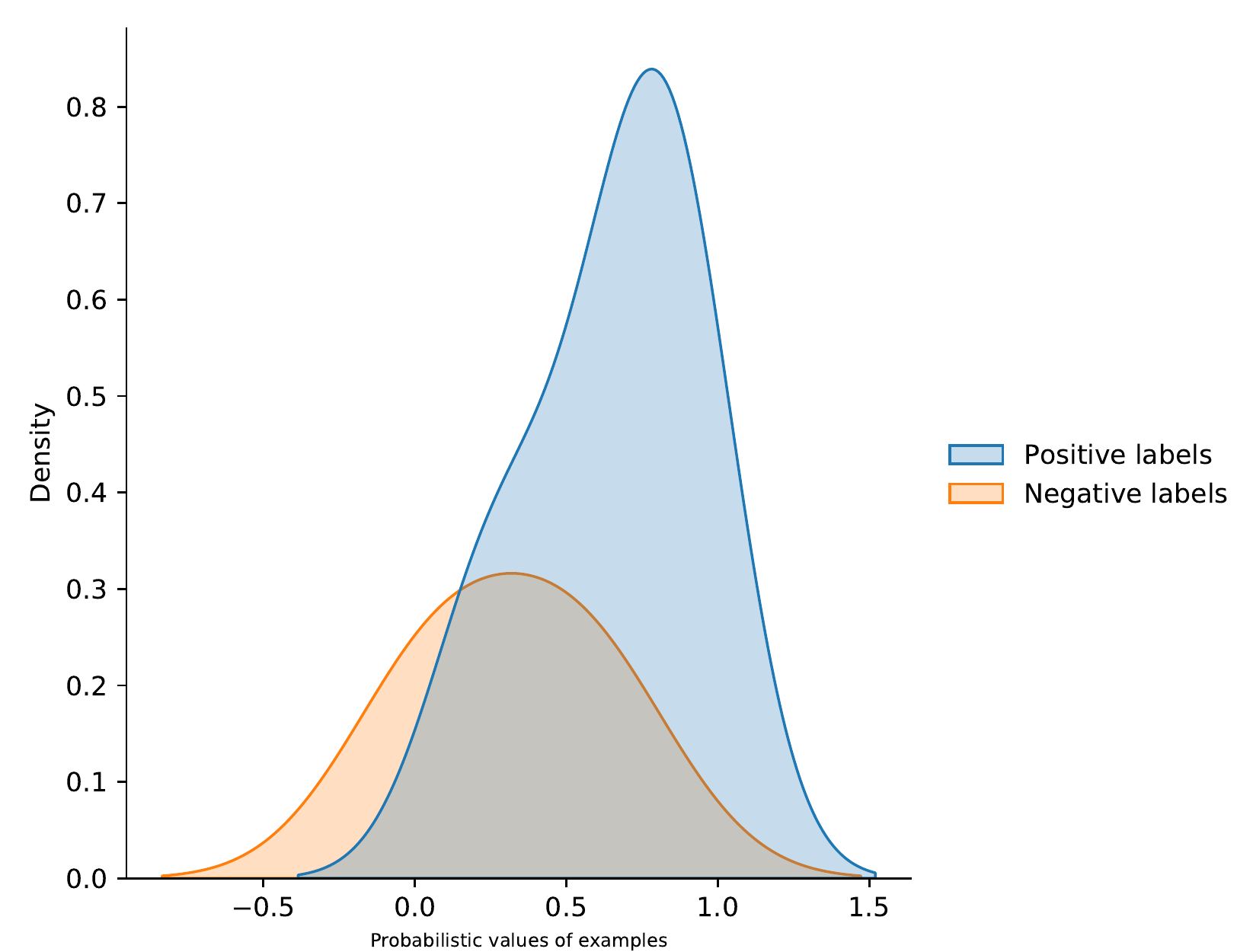}
        \caption{The distribution of examples in the \textit{predecessor} task when $\sigma = 3$.}
        \label{fig:dis_pre}
\end{figure}

\begin{figure}[H]
        \centering
        \includegraphics[width=\linewidth]{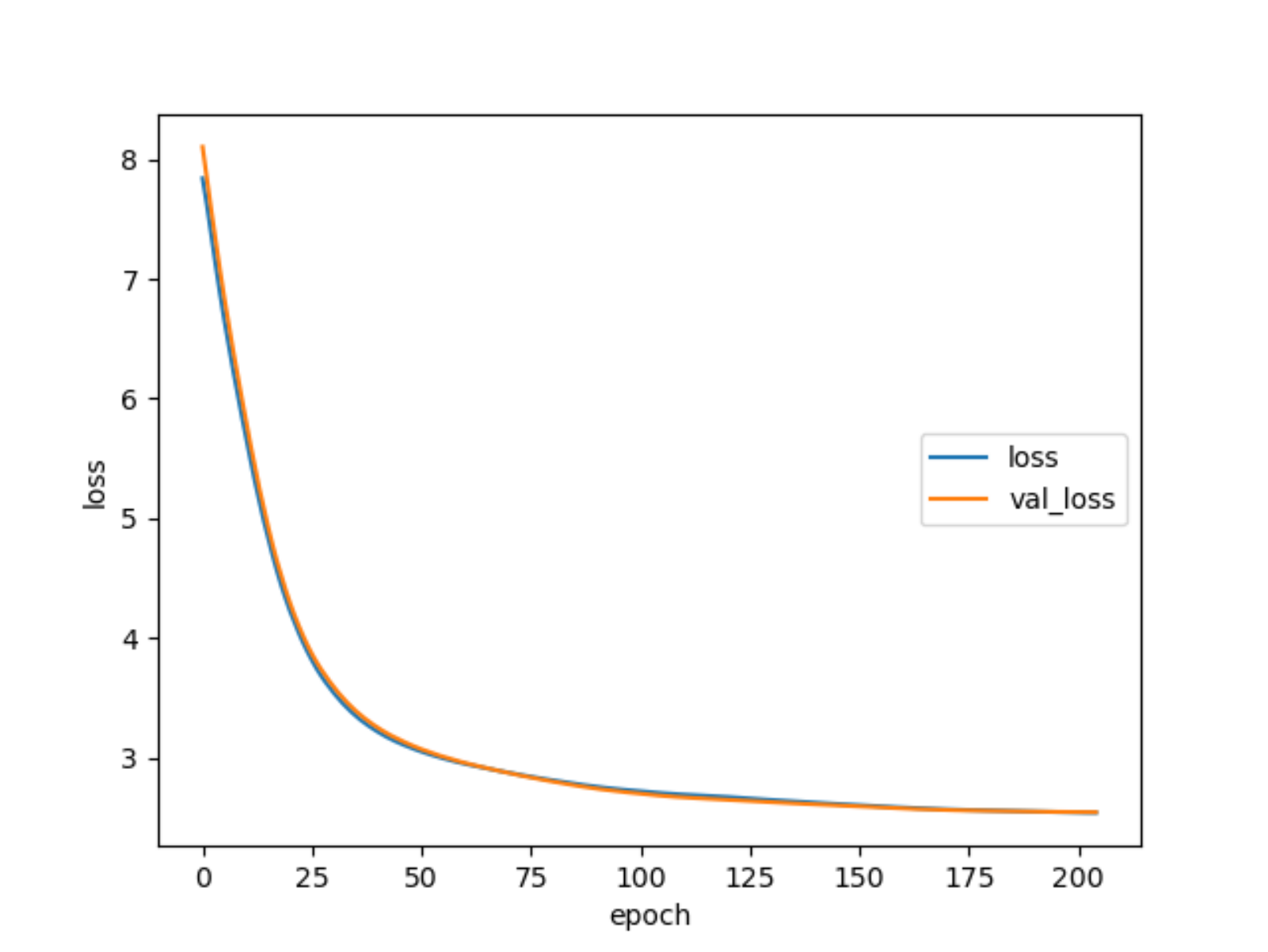}
        \caption{Learning curve in the \textit{lessthan} task.}
        \label{fig:loss_lt}
\end{figure}

\begin{figure}[H]
        \centering
        \includegraphics[width=\linewidth]{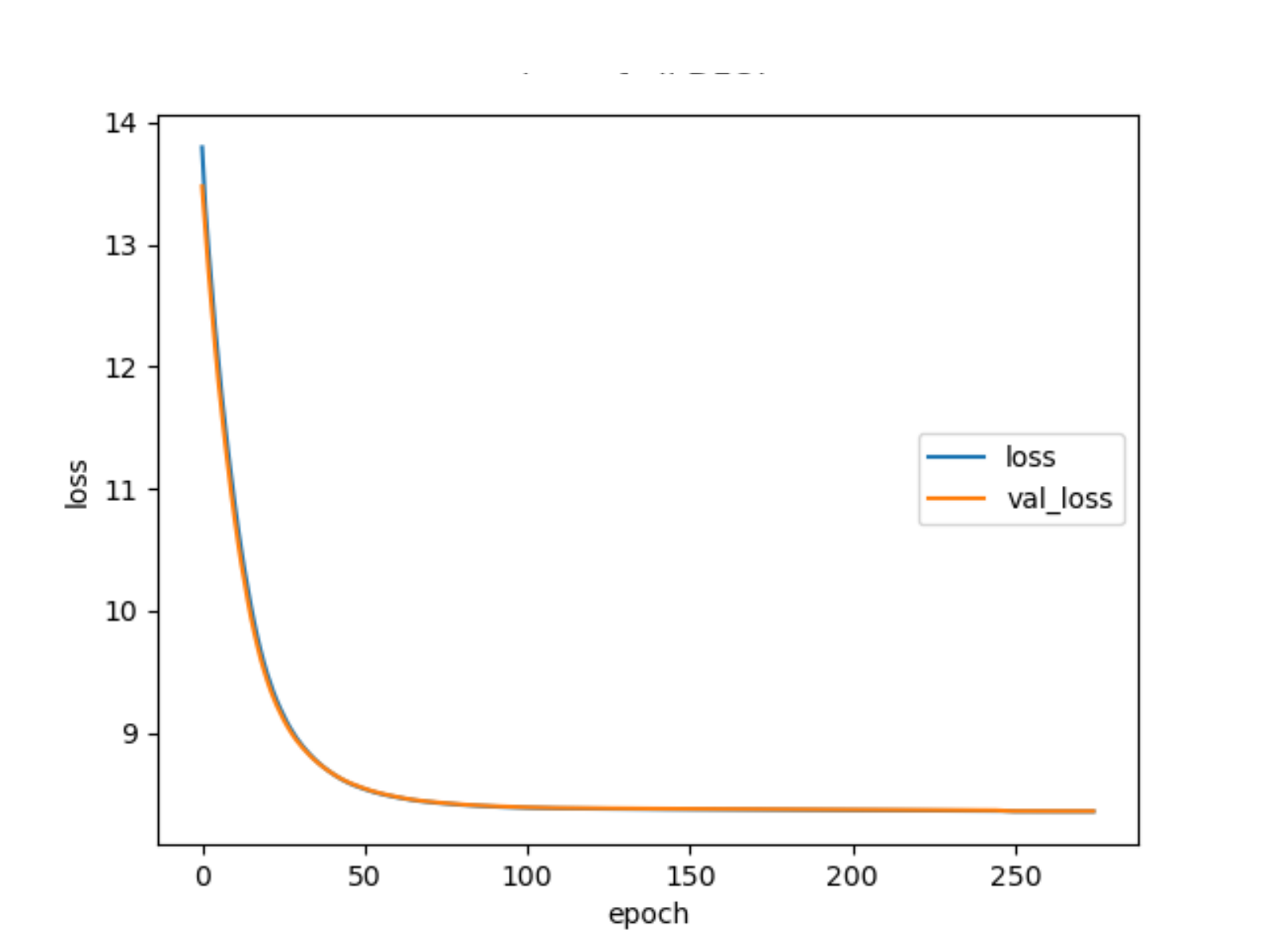}
        \caption{Learning curve in the \textit{predecessor} task.}
        \label{fig:loss_pre}
\end{figure}

Figures~\ref{fig:dis_less} and \ref{fig:dis_pre} present the dirtributions of the background assumptions, positive examples, and negative examples when $\sigma = 3$ in the \textit{lessthan} and \textit{predecessor} tasks. 
Figures~\ref{fig:loss_lt} and \ref{fig:loss_pre} present the learning curves in the \textit{lessthan} and \textit{predecessor} tasks.


\begin{table}[H]
    \caption[]{The descriptions of the knowledge bases.}
    \centering
    \label{datasetd}
    \begin{tabular}{lrrr}
    \toprule
    Dataset   & \#Object & \#Relation & \#Fact \\ 
    \midrule
    Countries & 252    & 2        & 1158 \\
    Nations   & 14     & 56       & 2565 \\
    UMLS      & 135    & 49       & 6529 \\ 
    \bottomrule
    \end{tabular}
\end{table}

\section{Experimental Results}\label{logic_program_app}

We present the results of comparing DFOL with the baselines in the classic ILP task in Section~\ref{com_app}. We also present symbolic logic programs generated by DFOL on classic ILP datasets and knowledge bases in Section~\ref{logic_app}.

\subsection{Comparison on Classic ILP Datasets}
\label{com_app}
\begin{table}[H]
    \caption {The results on ILP datasets. The symbols $\checkmark$, $\ast $ , and $-$ indicate that the accuracy of the generated logic program is equal to 100\%, less than 100\%, and equal to 0\%, respectively. }
    \label{comparing}
    \centering
    \begin{tabular}{llrrr}
    \toprule
    \small{Domain}     & \small{Task}          & \small{$\partial$ILP}  & \small{NeuralLP}   & \small{DFOL} \\ \midrule
    \small{Arithmetic} & \small{Predecessor}  & $\checkmark$  & $\checkmark$  & $\checkmark$ \\
     & \small{Odd}        & $\checkmark$ & $-$
        & $\checkmark$ \\
     &  \small{Even / Succ2  (10)}       & $\checkmark$  & $-$ 
     & $\checkmark$ \\
     & \small{Even / Succ2  (20)}       & $-$  & $-$ 
     & $\checkmark$ \\
     & \small{Lessthan}     & $\checkmark$  & $\checkmark$  & $\checkmark$ \\
     & \small{Fizz}                  & $\checkmark$  & $-$
                &$\checkmark$\\
     & \small{Buzz}                 & $\checkmark$ & $-$
          &$\checkmark$ \\
    \small{Lists}      & \small{Member}              & $\checkmark$  & $\ast $      & $\checkmark$  \\
          & \small{Length}              & $\checkmark$  & $-$
              &  $\checkmark$  \\
    \small{Family Tree}& \small{Son}                  & $\checkmark$  & $\ast $      & $\checkmark$ \\
    & \small{Grandparent}  & $\checkmark$  & $\checkmark$  & $\checkmark$ \\
    & \small{Husband}               & $\checkmark$  & $-$
       & $-$
        \\
    & \small{Uncle}               & $\checkmark$ & $-$
     &$-$
      \\
    & \small{Relatedness}          & $\checkmark$  & $\ast $     &  $\checkmark$ \\
    & \small{Father}                & $\checkmark$  & $-$
        &    $\checkmark$ \\
    \small{Graphs}     & \small{Directed Edge}           & $\checkmark$  &  $\ast $      &  $\checkmark$ \\
    & \small{Adjacent to Red}           & $\checkmark$  & $-$
        &  $\checkmark$ \\
         & \small{Two Children}            & $\checkmark$  & $-$
           &  $\checkmark$  \\
         & \small{Graph Coloring (6)}           & $\checkmark$ & $-$
             &   $\checkmark$ \\

& \small{Graph Coloring (10)}           & $-$ & $-$
                  &   $\checkmark$ \\

         & \small{Connectedness}               & $\checkmark$ & $\ast $    &   $\checkmark$\\
         & \small{Cyclic}                     & $\checkmark$  & $-$
            &  $\checkmark$ \\ \bottomrule
    \end{tabular}
    \end{table}

\subsection{Generated Symbolic Logic Programs}
\label{logic_app}
\begin{enumerate}
    \item \textbf{The predecessor task:}  
    The entity set is:
    \begin{align*}
        E = \{ 0,1, \dots,8 \}
    \end{align*}
    The training background assumption set is:
    \begin{align*}
        \mathcal{B} = \{ succ(X,X+1) \mid  X \in E \} \cup {zero(0)}
    \end{align*}
    The training positive example set is: 
    \begin{align*}
        \mathcal{P} = \{ pre(X+1,X) \mid X\in E \}
    \end{align*}
    For the test data, we use positive and negative examples of the \textit{pre} predicate on numbers greater than 9.
    The result is shown as follows:
    \begin{figure}[H]
        \includegraphics[width=.55\linewidth, center]{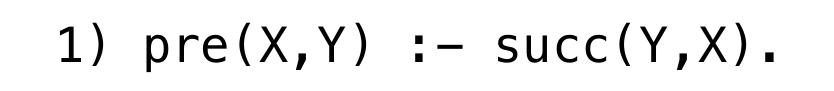}
    \end{figure}

    \item \textbf{The odd task:} 
    The entity set is:
    \begin{align*}
        E = \{ 0,1, \dots,8 \}
    \end{align*}
    The training background assumption set is:
    \begin{align*}
        \mathcal{B} = \{ succ(X,X+1) \mid   X \in E \} \cup {zero(0)}
    \end{align*}
    The training positive example set is: 
    \begin{align*}
        \mathcal{P} = \{ odd(X) \mid X \ \text{mod}  \ 2 = 1 ,X\in E \cup \{9\}\}
    \end{align*}
    For the test data, we use positive and negative examples of the \textit{odd} predicate on numbers greater than 9.
    The result is shown as follows:
    \begin{figure}[H]
        \includegraphics[width=.8\linewidth, center]{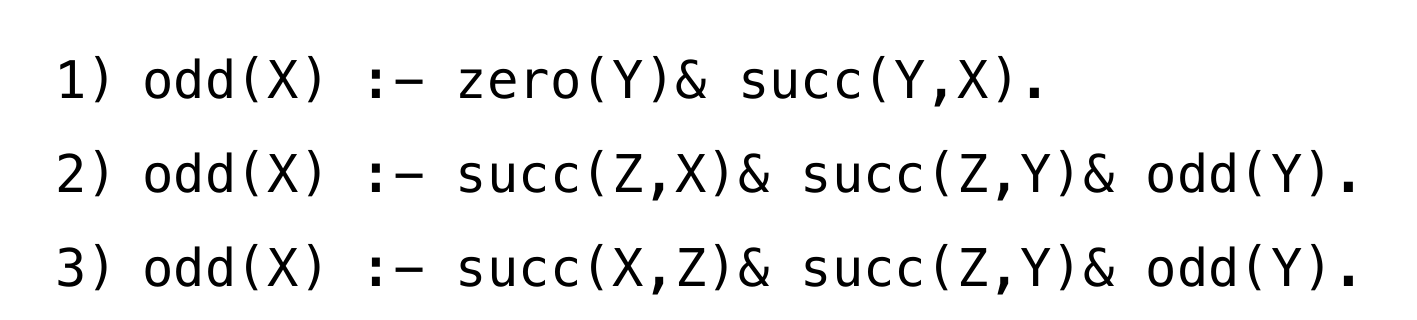}
    \end{figure}

    \item \textbf{The even (10) task:}
    The entity set is:
    \begin{align*}
        E = \{ 0,1, \dots,8\}
    \end{align*}
    The training background assumption set is:
    \begin{align*}
        \mathcal{B} = \{ succ(X,X+1) \mid  X \in E  \} \cup {zero(0)}
    \end{align*}
    The training positive example set is: 
    \begin{align*}
        \mathcal{P} = \{ even(X) \mid X \ \text{mod}  \ 2 = 0 ,X\in E \cup \{9\}  \}
    \end{align*}
    For the test data, we use positive and negative examples of the \textit{even} predicate on numbers greater than 9.
    The result is shown as follows:
    \begin{figure}[H]
        \includegraphics[width=0.8\linewidth, center]{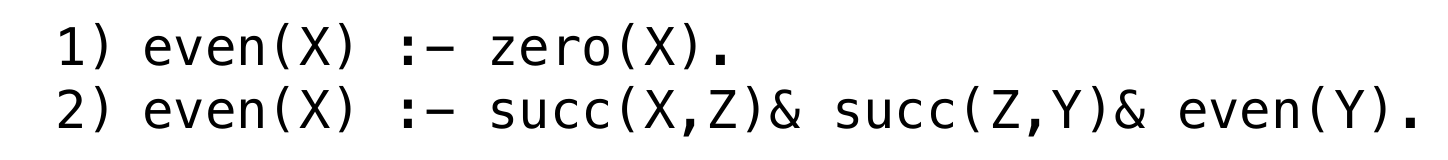}
    \end{figure}

    \item \textbf{The even (20) task:}
    The entity set is:
    \begin{align*}
        E = \{ 0,1, \dots,18\}
    \end{align*}
    The training background assumption set is:
    \begin{align*}
        \mathcal{B} = \{ succ(X,X+1) \mid  X \in E \} \cup {zero(0)}
    \end{align*}
    The training positive example set is: 
    \begin{align*}
        \mathcal{P} = \{ even(X) \mid X \ \text{mod}  \ 2 = 0 ,X\in E \cup \{19\}   \}
    \end{align*}
    For the test data, we use positive and negative examples of the \textit{even} predicate on numbers greater than 19.
    The result is shown as follows:
    \begin{figure}[H]
        \includegraphics[width=0.8\linewidth, center]{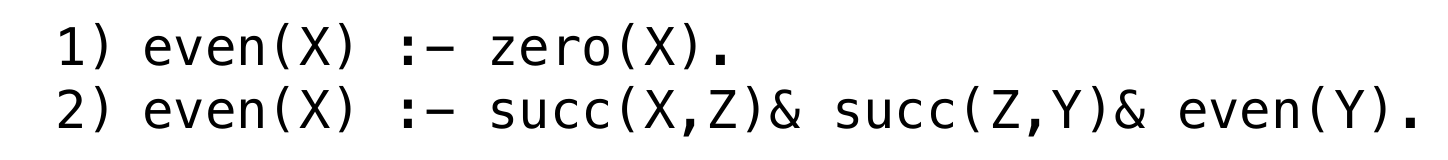}
    \end{figure}

    \item \textbf{The succ2 task:}
    The entity set is:
    \begin{align*}
        E = \{ 0,1, \dots,9 \}
    \end{align*}
    The training background assumption set is:
    \begin{align*}
        \mathcal{B} = \{ succ(X,X+1) \mid  X \in E-\{9\} \} \cup {zero(0)}
    \end{align*}
    The training positive example set is: 
    \begin{align*}
        \mathcal{P} = \{ succ2(X,X+2) \mid X\in E-\{ 8,9 \}\}
    \end{align*}
    For the test data, we use positive and negative examples of the \textit{succ2} predicate on numbers greater than 9.
    The result is shown as follows:
\begin{figure}[H]
    \includegraphics[width=.75\linewidth, center]{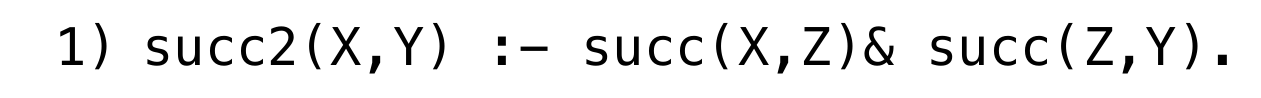}
\end{figure}

    \item \textbf{The {lessthan} task:}
    The entity set is:
    \begin{align*}
        E = \{ 0,1, \dots,9 \}
    \end{align*}
    The training background assumption set is:
    \begin{align*}
        \mathcal{B} = \{ succ(X,X+1) \mid  X \in E-\{9\} \} \cup {zero(0)}
    \end{align*}
    The training positive example set is: 
    \begin{align*}
        \mathcal{P} = \{ lt(X,Y) \mid X < Y ,X\in E, Y\in E\}
    \end{align*}
    For the test data, we use positive and negative examples of the \textit{lt} predicate on numbers greater than 9.
    The result is shown as follows:
    \begin{figure}[H]
        \centering
        \includegraphics[width=.7\linewidth]{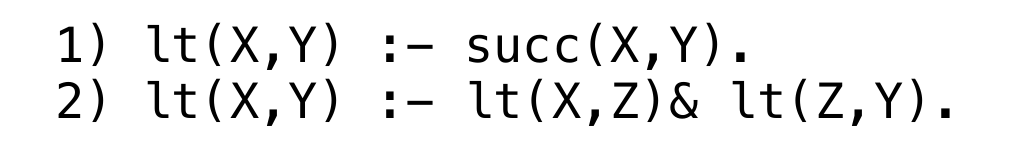}
    \end{figure}

\item  \textbf{The {fizz} task:}
The entity set is:
\begin{align*}
    E = \{ 0,1,2,3,4,5,6 \}
\end{align*}
The training background assumption set is:
\begin{align*}
    \mathcal{B} = \{ succ(X,X+1) \mid X \in E-\{6\} \} \cup {zero(0)}
\end{align*}
The training positive example set is: 
\begin{align*}
    \mathcal{P} = \{ fizz(X) \mid X \ \text{mod} \ 3 =0  ,X\in E\}
\end{align*}
For the test data, we use positive and negative examples of the \textit{fizz} predicate on numbers greater than 6.
The result is shown as follows:
\begin{figure}[H]\
    \centering
    \includegraphics[width=.8\linewidth]{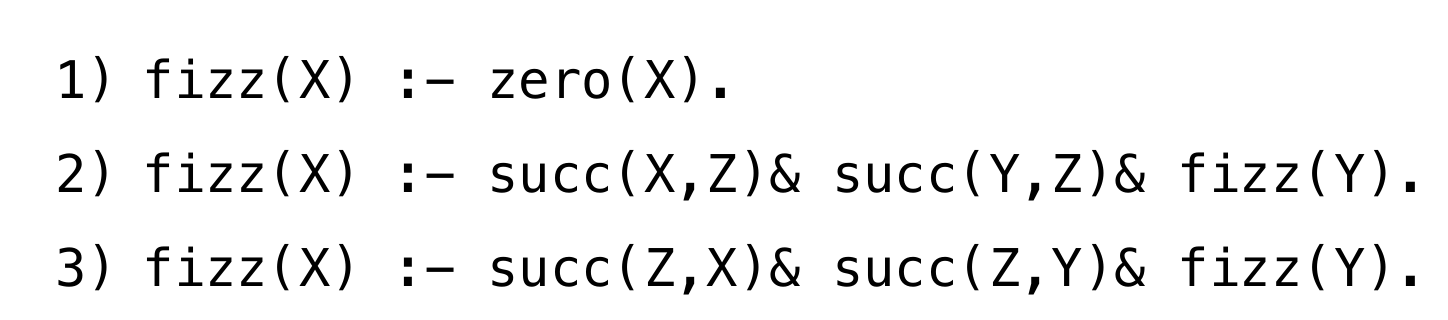}
\end{figure}

\item  \textbf{The {buzz} task:}
The entity set is:
\begin{align*}
    E = \{ 0,1,2,3,4,5,6,7,8,9 \}
\end{align*}
The training background assumption set is:
\begin{align*}
    \mathcal{B} = \{ & succ(X,X+1) \mid X \in E-\{9\} \} \cup \{zero(0)\} \cup \\& \{pred3(X, X+3) \mid X \in E-\{7,8,9\} \} \cup \\& \{ pred2(X, X + 2) \mid X \in E-\{8,9\} \}
\end{align*}
The training positive example set is: 
\begin{align*}
    \mathcal{P} = \{ buzz(X) \mid X \ \text{mod} \ 5 =0  ,X\in E\}
\end{align*}
For the test data, we use positive and negative examples of the \textit{buzz} predicate on numbers greater than 9.
The result is shown as follows:
\begin{figure}[H]
    \centering
    \includegraphics[width=.8\linewidth]{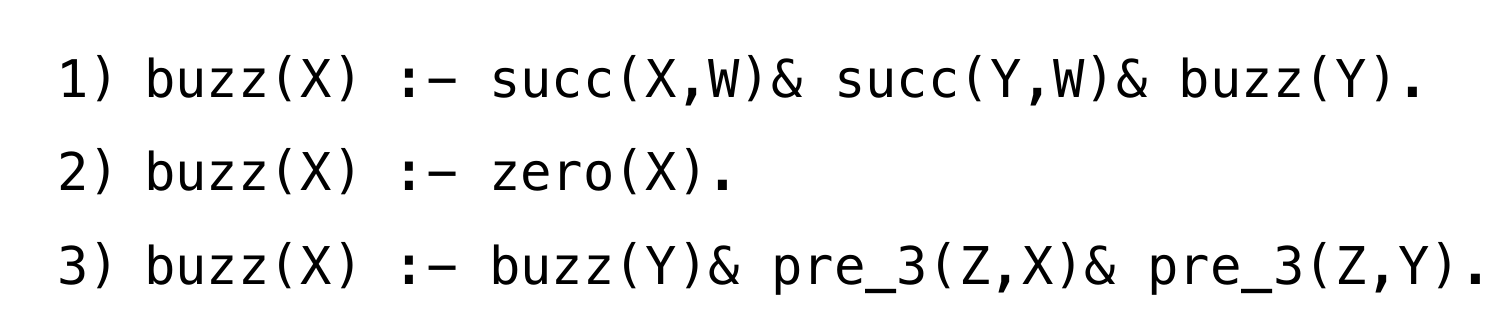}
\end{figure}

\item  \textbf{The member task:}
The task is to learn the \textit{member} relation on lists, where $member(X,Y)$ if $X$ is an element in list $Y$. 
Besides, $cons(X, Y)$ if the node after $X$ is node $Y$, and $value(X, Y) $if the value of node $X$ is $Y$.

The training background assumption set is:
\begin{align*}
    \mathcal{B} = \{ & cons([4,3,2,1],[3,2,1]), cons([3,2,1],[2,1]),\\&  cons([2,1],[1]), value([4,3,2,1],4),  value([1],1), \\&  value([3,2,1],3),  value([2,1],2)\}
\end{align*}

The positive example set is:
\begin{align*}
    \mathcal{P} = \{ &member(4,[4,3,2,1]),
    member(3,[4,3,2,1]),
\\&    member(3,[3,2,1]),
    member(2,[4,3,2,1]),
\\&    member(2,[3,2,1]),
    member(2,[2,1]),
\\&    member(1,[4,3,2,1]),
    member(1,[3,2,1]),
\\&    member(1,[2,1]),
    member(1,[1])
    \}
\end{align*}

The result is shown as follows:
\begin{figure}[H]
    \centering
    \includegraphics[width=.9\linewidth]{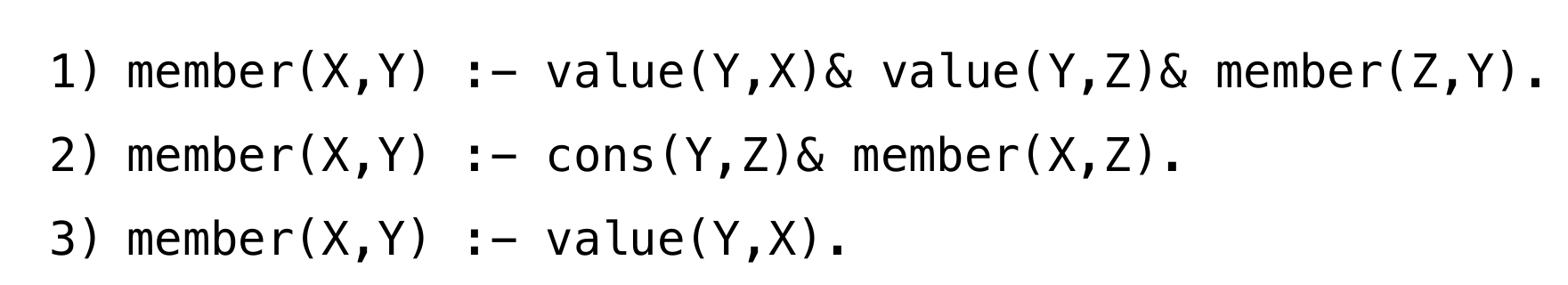}
\end{figure}

\item  \textbf{The length task:}
The task is to learn the \textit{length} relation and $length(X, Y)$ is true if the length of list $X$ is $Y$.

The training background assumption set consists of $con$, $succ$, and $zero$ relations.

The background assumption set is:
\begin{align*}
    \mathcal{P} = \{ &succ(0,1), succ(1,2), succ(2,3),  cons([2,1],[1], \\& cons([4,3,2,1],[3,2,1]), cons([3,2,1],[2,1]))
    \}
\end{align*}

The positive example set is:
\begin{align*}
    \mathcal{P} = \{& length([4,3,2,1],4), length([3,2,1], 3),\\& length([2,1], 2)\}
\end{align*}

The result is shown as follows:
\begin{figure}[H]
    \centering 
    \includegraphics[width=.9\linewidth]{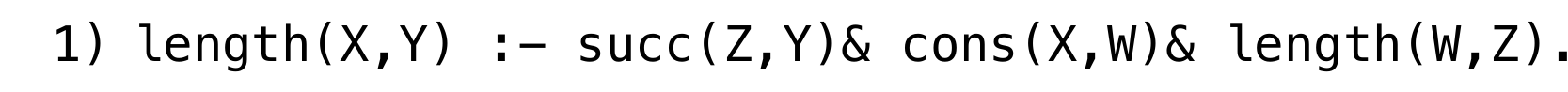}
\end{figure}

\item  \textbf{The son task:}
This task is to learn $son$ relation based on $father$, $sister$, and $brother$ relations.

The background assumption set is:
\begin{align*}
    \mathcal{B} = \{&father(a, b), father(a, c), father(d, e), \\&father(d, f), father(g, h), father(g, i), \\&brother(b, c), brother(c, b), brother(e, f), \\&sister(f, e), sister(h, i), sister(i, h)\}
\end{align*}

The positive example set is:
\begin{align*}
    \mathcal{P} = \{ son(b, a), son(c, a), son(e, d)
    \}
\end{align*}

The result is shown as follows:
\begin{figure}[H]
    \centering
    \includegraphics[width=.7\linewidth]{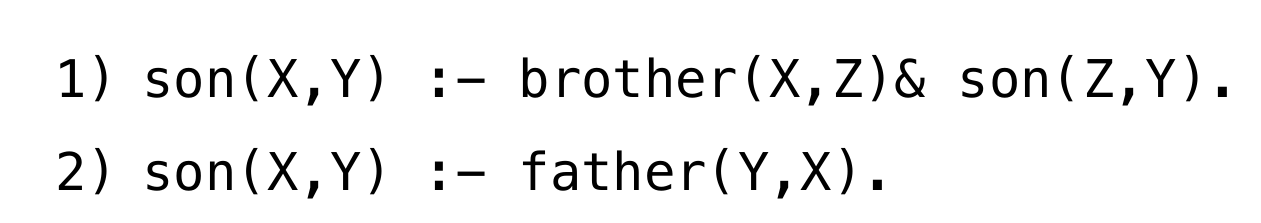}
\end{figure}

\item  \textbf{The grandparent task:}
The correct generated logic program is presented in Section~\ref{learnILP}.

\item  \textbf{The relatedness task:}
The task is to learn the $relatedness$ relation from facts about family relations involving the $parent$ relation.

The background assumption set is:
\begin{align*}
    \mathcal{B} = \{&parent(a, b), parent(a, c), parent(c, e), \\& parent(c, f), parent(d, c), parent(g, h) \},
\end{align*}

The positive example set is:
\begin{align*}
    \mathcal{P} = \{ &relatedness(a, b), relatedness(a, c), \\& relatedness(a, e), relatedness(a, f), \\ & relatedness(f, a), relatedness(a, a), \\& relatedness(d, b), relatedness(h, g)
    \}
\end{align*}

The result is shown as follows:

\begin{figure}[H]
    \includegraphics[width=\linewidth, center ]{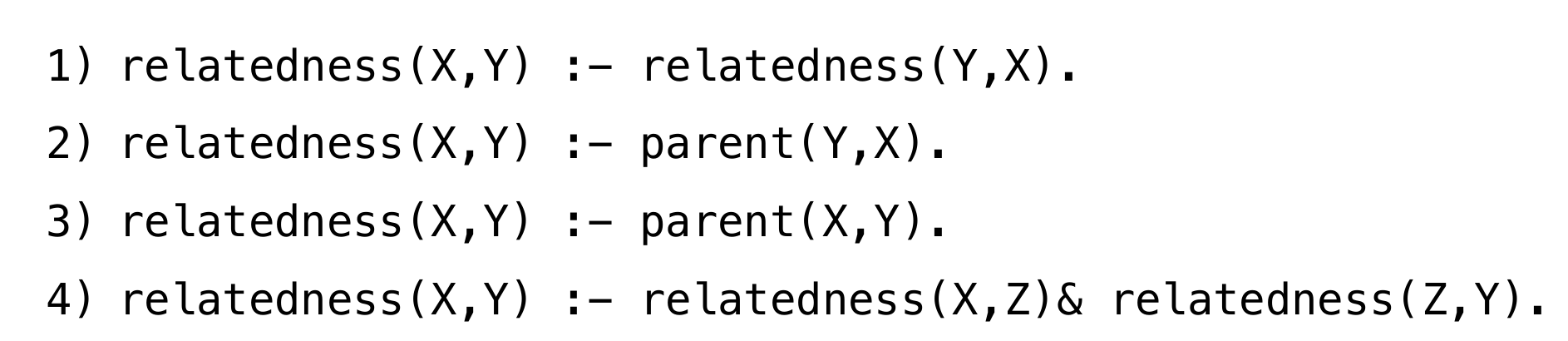}
\end{figure}

\item  \textbf{The father task:}
The task is to learn the $father$ relation given background assumptions as follows:
\begin{figure}[H]\
    \centering
    \includegraphics[width=.9\linewidth]{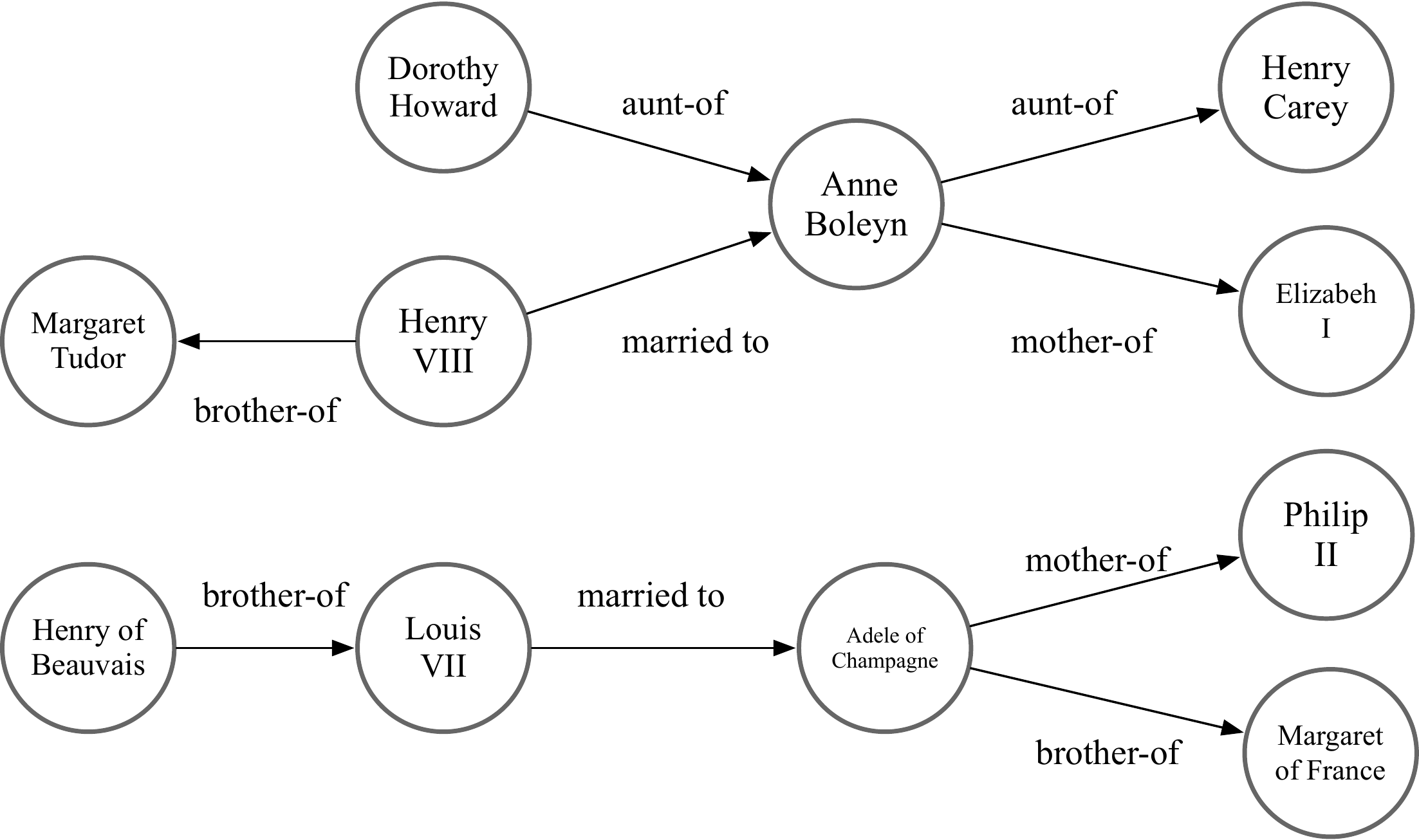}
\end{figure}
Moreover, two positive examples are:
\begin{align*}
    \mathcal{P} = \{ &father({Louis VII},{Phillip II}),\\& father({Henry VIII},{Elizabeth I})
    \}
\end{align*}

The result is shown as follows:

\begin{figure}[H]
    \includegraphics[width=0.9\linewidth]{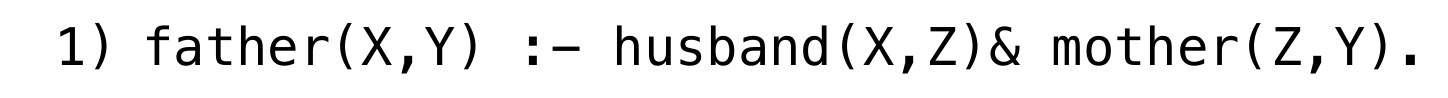}
\end{figure}

\item  \textbf{The directed edge (\textit{d-edge}) task:}
The task is to learn the $d-edge$ relation, which is true of $X$ and $Y$ if there is an edge, in either direction, between $X$ and $Y$.

The background assumption set is:
\begin{align*}
    \mathcal{B} = \{{edge(a, b), edge(b, d), edge(c, c)} \},
\end{align*}

The positive example set is:
\begin{align*}
    \mathcal{P} = \{&  \textit{d-edge}(a, b), \textit{d-edge}(b, a), \textit{d-edge}(b, d),\\& \textit{d-edge}(d, b), \textit{d-edge}(c,c)
    \}
\end{align*}
The result is shown as follows:
\begin{figure}[H]
    \centering
    \includegraphics[width=.55\linewidth]{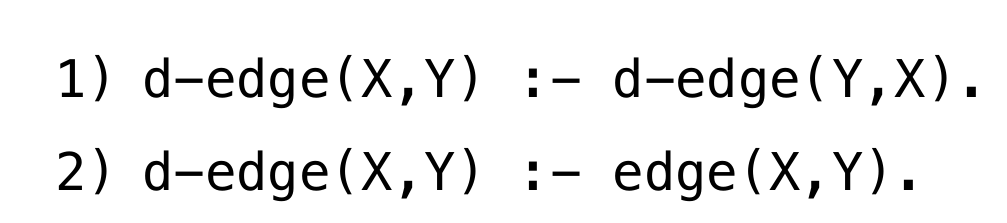}
\end{figure}

\item  \textbf{The adjacent to red (\textit{ared}) task:}
The task is to learn the predicate is adjacent to a red node.
The background assumption set is:
\begin{align*}
    \mathcal{B} = \{ &edge(a, b), edge(b, a), edge(c, d), edge(c, e),\\& edge(d, e), color(a, red), color(b, green), \\& color(c, red),   color(d, red), color(e, green) \},
\end{align*}
The positive example set is:
\begin{align*}
    \mathcal{P} = \{& ared(b), ared(c), ared(e)
    \}
\end{align*}
The result is shown as follows:
\begin{figure}[H]
    \centering
    \includegraphics[width=.7\linewidth]{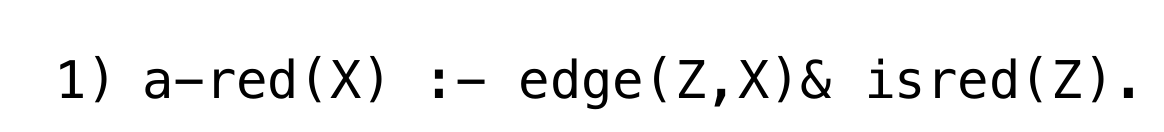}
\end{figure}

\item  \textbf{The two children (\textit{tc}) task:}
The task is to learn the predicate has at least two children.
The background assumption set is:
\begin{align*}
    \mathcal{B} = \{ &edge(a, b), edge(a, c), edge(b, d), edge(c, d),\\& edge(c, e), edge(d, e) \} \cup \{neq(X,Y) \mid X \neq Y\}
\end{align*}

The positive example set is:
\begin{align*}
    \mathcal{P} = \{& tc(a), tc(c)
    \}
\end{align*}
The result is shown as follows:

\begin{figure}[H]
    \centering
    \includegraphics[width=.8\linewidth]{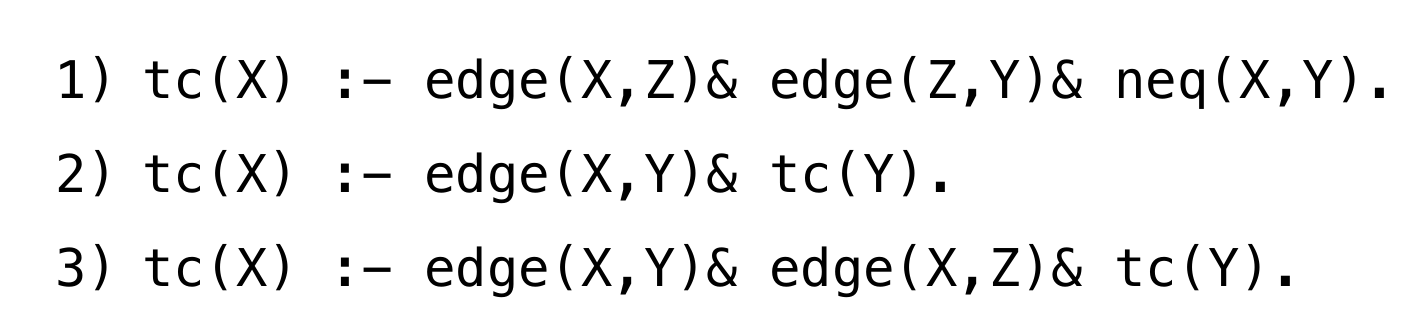}
\end{figure}

\item  \textbf{The graph coloring (\textit{gc}) (6) task:}
The task here is to learn the $gc$ predicate that is true of a node $X$ if $X$ is adjacent to a node of the same color. There are six nodes in this task. 
The background assumption set is:
\begin{align*}
    \mathcal{B} = \{ &edge(a, b), edge(b, c), edge(b, d), edge(c, e),\\& edge(e, f),color(a, green), color(b, red), \\& color(c, green), color(d, green), \\& color(e, red), color(f, red) \}
\end{align*}

The positive example set is:
\begin{align*}
    \mathcal{P} = \{& gc(e)
    \}
\end{align*}
The result is shown as follows:
\begin{figure}[H]
    \centering
    \includegraphics[width=.8\linewidth]{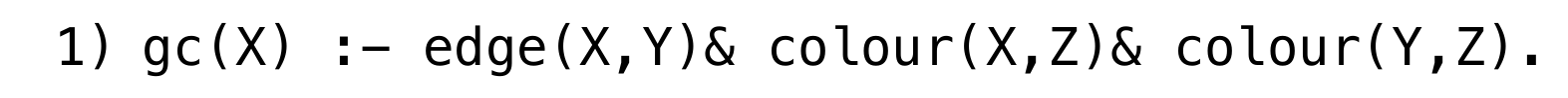}
\end{figure}

\item  \textbf{The \textit{gc} (10) task:}
The task here is to learn the $gc$ predicate that is true of a node $X$ if $X$ is adjacent to a node of the same color. There are 10 nodes in this task. 
The background assumption set is:
\begin{align*}
    \mathcal{B} = \{ & edge(a,b),
    edge(b,c),
    edge(b,d),
    edge(c,e),
    \\& edge(e,f),
    edge(a,g),
    edge(d,h),
    edge(b,i),
    \\&edge(b,m),
     color(a,green),
color(b,red),
\\& color(c,green),
 color(d,green),
color(e,red),
\\& color(f,red),
 color(g,green),
color(h,green),
\\& color(i,red),
color(m,green),
color(f,red) \}
\end{align*}

The positive example set is:
\begin{align*}
    \mathcal{P} = \{
    gc(e),
    gc(f),
    gc(a),
    gc(g),
    gc(d),
    gc(h),
    gc(i),
    gc(b)
    \}
\end{align*}
The result is shown as follows:
\begin{figure}[H]
    \centering
    \includegraphics[width=.8\linewidth]{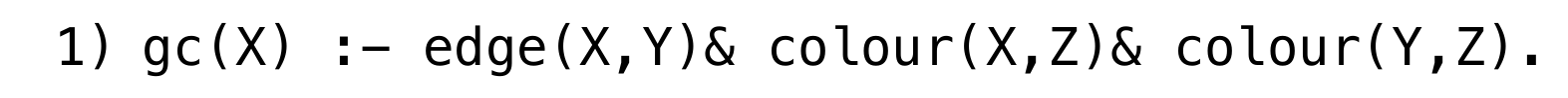}
\end{figure}

\item  \textbf{The connectedness task:}
The task is to learn the $connected(X, Y)$ relation that is true if there is some sequence of edge transitions connecting $X$ and $Y$.
The background assumption set is:
\begin{align*}
    \mathcal{B} = \{ edge(a, b), edge(b, c), edge(c, d), edge(b, a) \}
\end{align*}

The positive example set is:
\begin{align*}
    \mathcal{P} = \{& connectedness(a, b), connectedness(b, c), \\&connectedness(c, d), connectedness(b, a), \\& connectedness(a, c), connectedness(a, d), \\& connectedness(a, a), connectedness(b, d), \\& connectedness(b, a), connectedness(b, b)
    \}
\end{align*}
The result is shown as follows:

\begin{figure}[H]
    \centering
    \includegraphics[width=.9\linewidth]{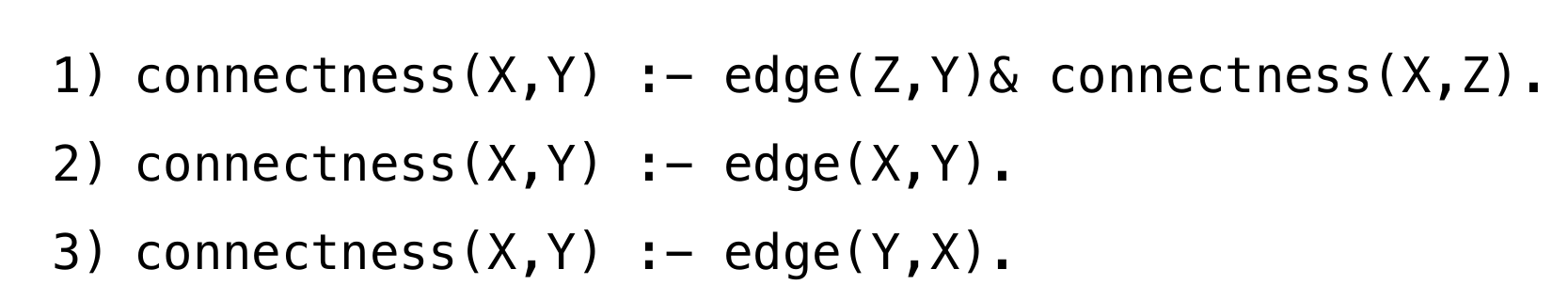}
\end{figure}

\item  \textbf{The cyclic task:}
The task is to learn the $cyclic$ predicate. This predicate is true of a node if there is a path, a sequence of edge connections, from that node back to itself. 
The background assumption set is:
\begin{align*}
    \mathcal{B} = \{& edge(a,b),
    edge(b,c),
    edge(c,a),
    edge(b,d),
    \\&
    edge(d,e),
    edge(d,f),
    edge(e,f),
    edge(f,e)\}
\end{align*}

The positive example set is:
\begin{align*}
    \mathcal{P} = \{&cyclic(a,a),
    cyclic(b,b),
    cyclic(c,c),
    cyclic(e,e),
    \\&
    cyclic(f,f),
    \}
\end{align*}
The result is shown as follows:

\begin{figure}[H]
    \centering
    \includegraphics[width=1\linewidth]{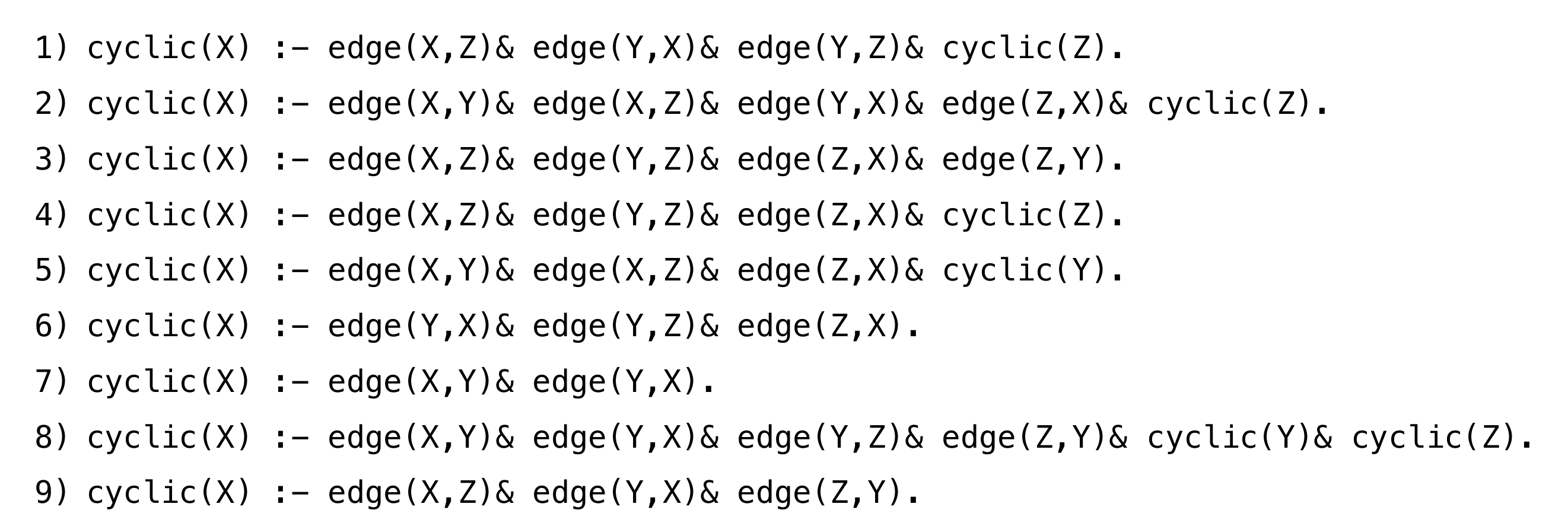}
\end{figure}

\item  \textbf{The Countries S1 task:}
The logic program in the Countries S1 knowledge base is:
\begin{figure}[H]
    \includegraphics[width=\linewidth]{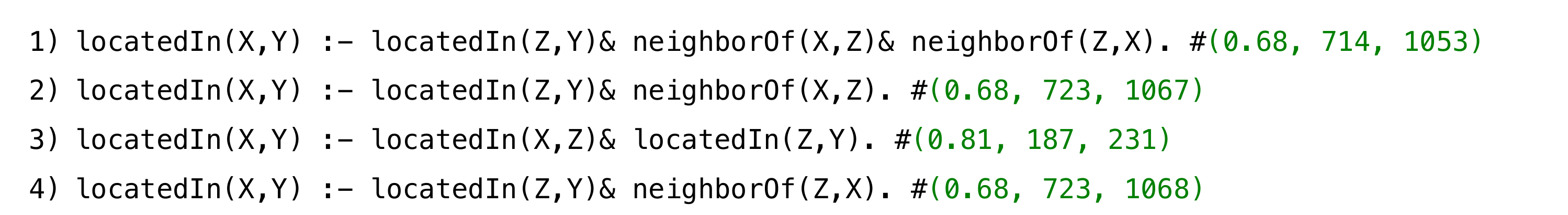}
\end{figure}

\item  \textbf{The Countries S2 task:}
The logic program in the Countries S2 knowledge base is:
\begin{figure}[H]
    \includegraphics[width=\linewidth]{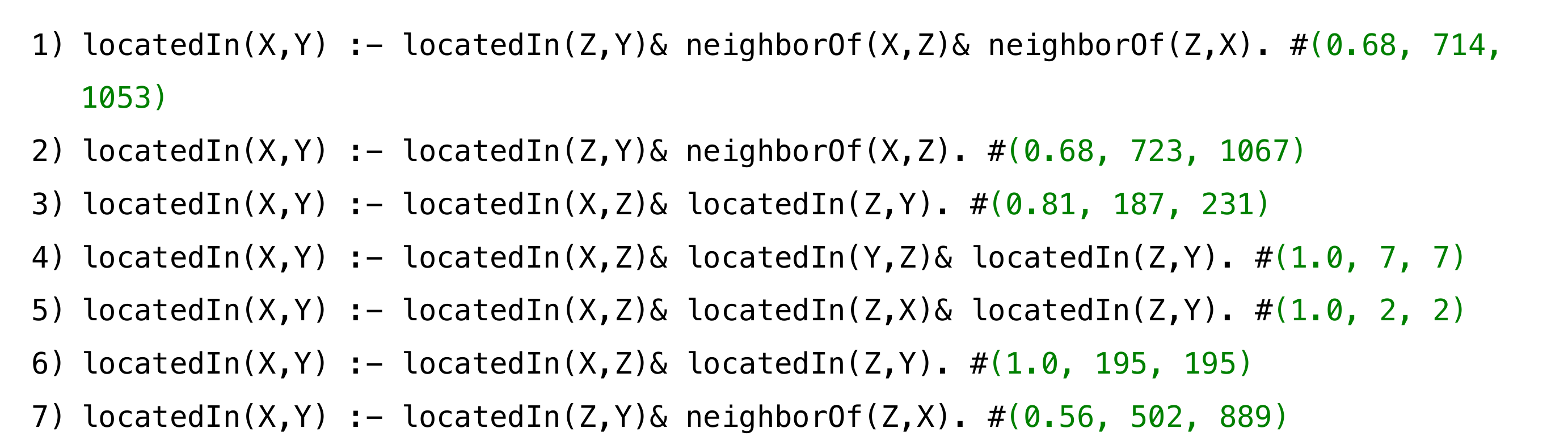}
\end{figure}

\item  \textbf{The Countries S3 task:}
The logic program in the Countries S3 knowledge base is:
\begin{figure}[H]
    \includegraphics[width=\linewidth]{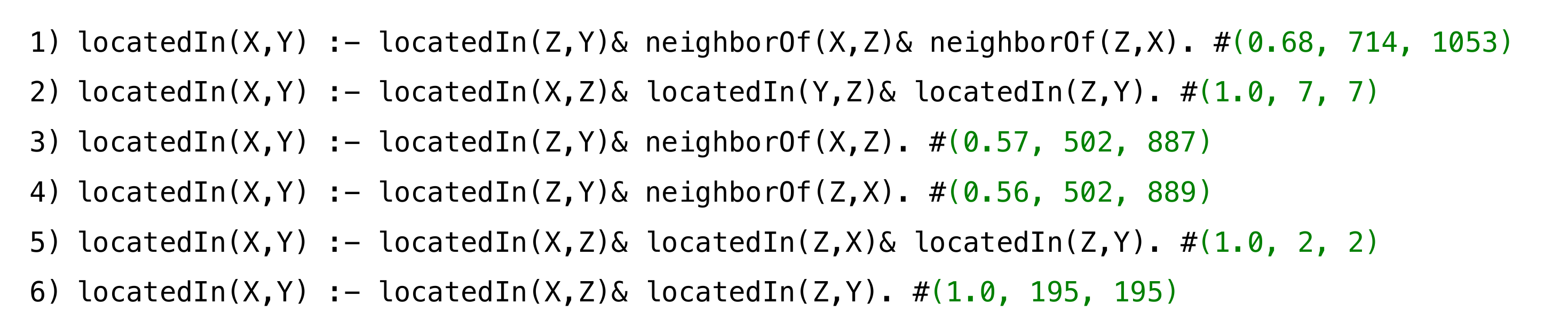}
\end{figure}

\item  \textbf{The Nations-blockpositionindex task:}
The logic program in the Nations knowledge base with the target atom $blockpositionindex$$(X,Y)$ is shown as follows:
\begin{figure}[H]
    \includegraphics[width=\linewidth]{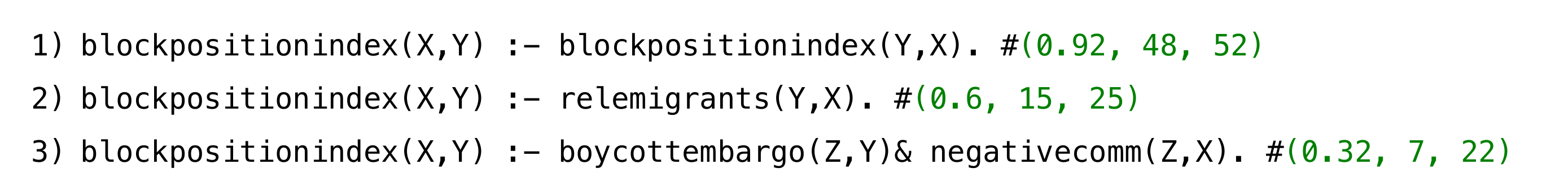}
\end{figure}

\item  \textbf{The Nations-negativecomm task:}   
The logic program in the Nations knowledge base  with the target atom $negativecomm(X,Y)$ is shown as follows:
\begin{figure}[H]
    \includegraphics[width=1\linewidth]{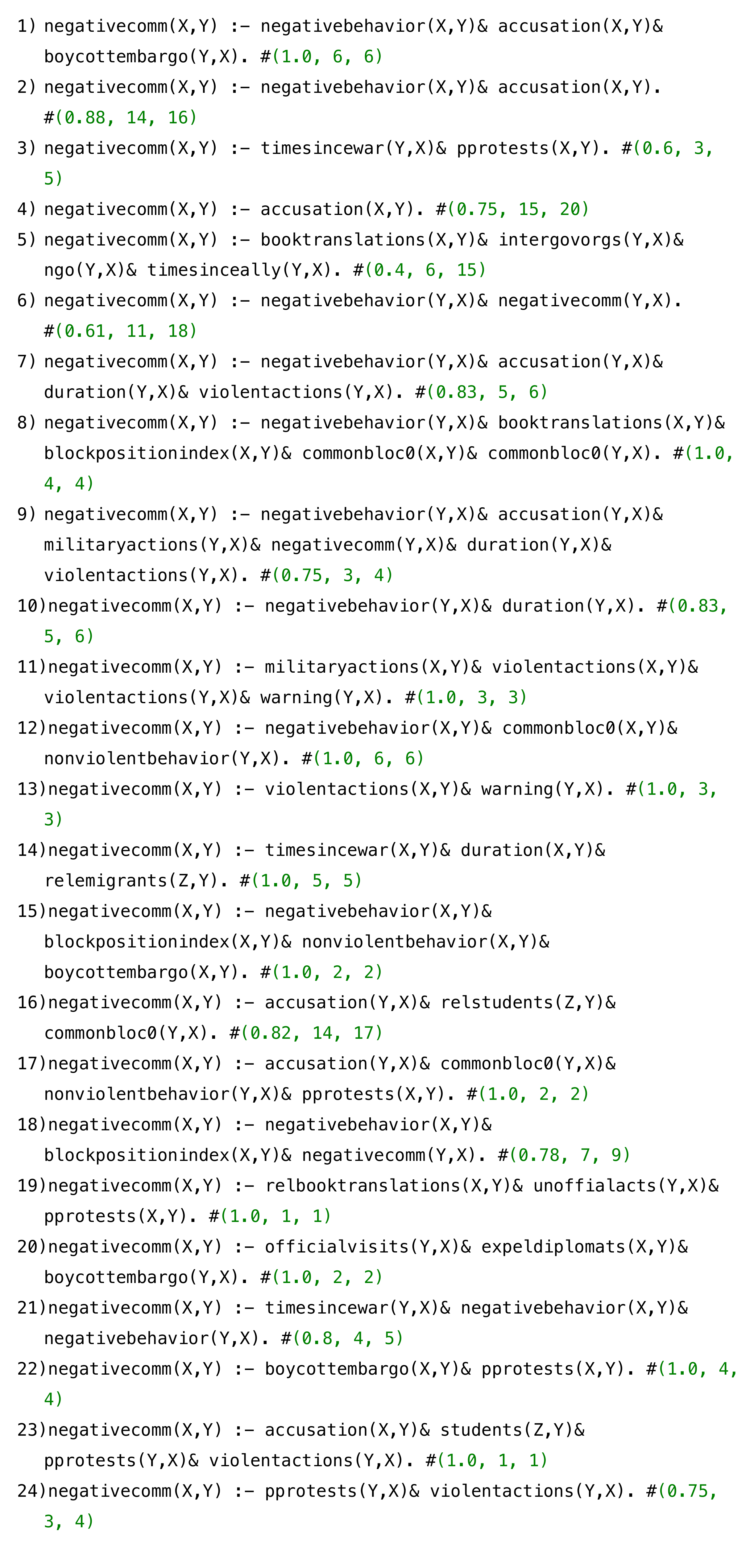}
\end{figure}

\item  \textbf{The UMLS-interacts\_with task:}
The logic program in the UMLS knowledge base  with the target atom $interacts\_with(X,Y)$ is shown as follows:
\begin{figure}[H]
    \includegraphics[width=1\linewidth]{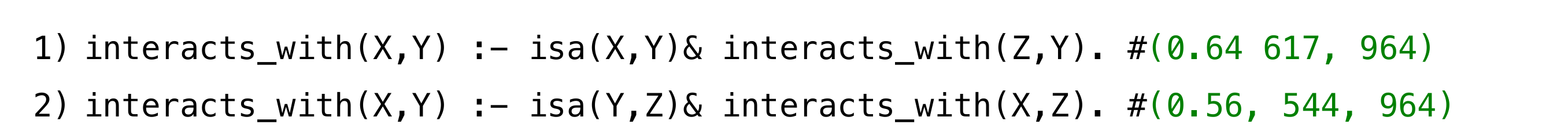}
\end{figure}

\item  \textbf{The Nations-intergovorgs3 task:}
The logic program in the Nations knowledge base  with the target atom $intergovorgs3(X,Y)$ is shown as follows:
\begin{figure}[H]
    \includegraphics[width=1\linewidth]{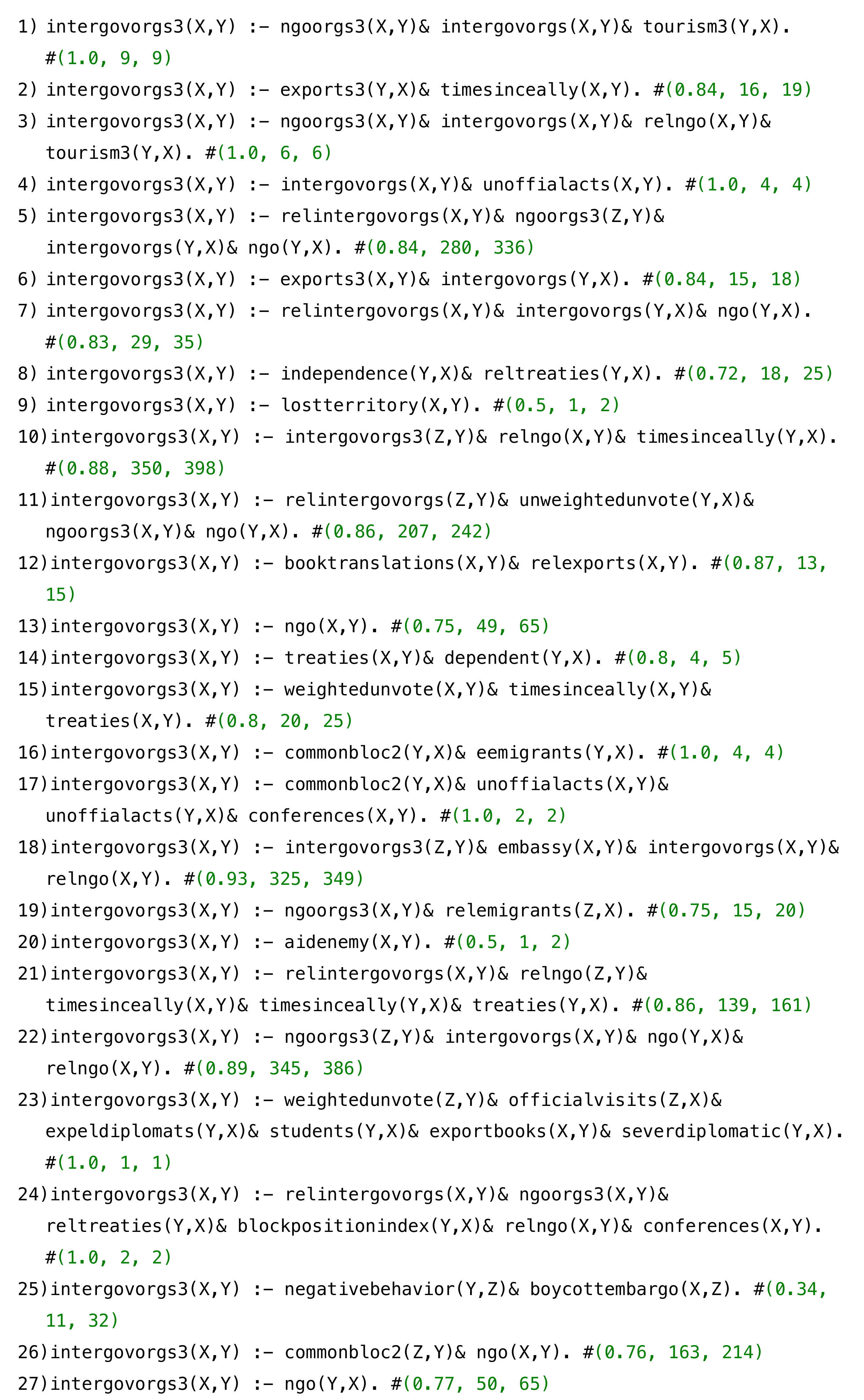}
\end{figure}

\item  \textbf{The UMLS-isa task:}
The logic program in the UMLS knowledge base  with the target atom $isa(X,Y)$ is shown as follows:
\begin{figure}[H]
    \includegraphics[width=\linewidth]{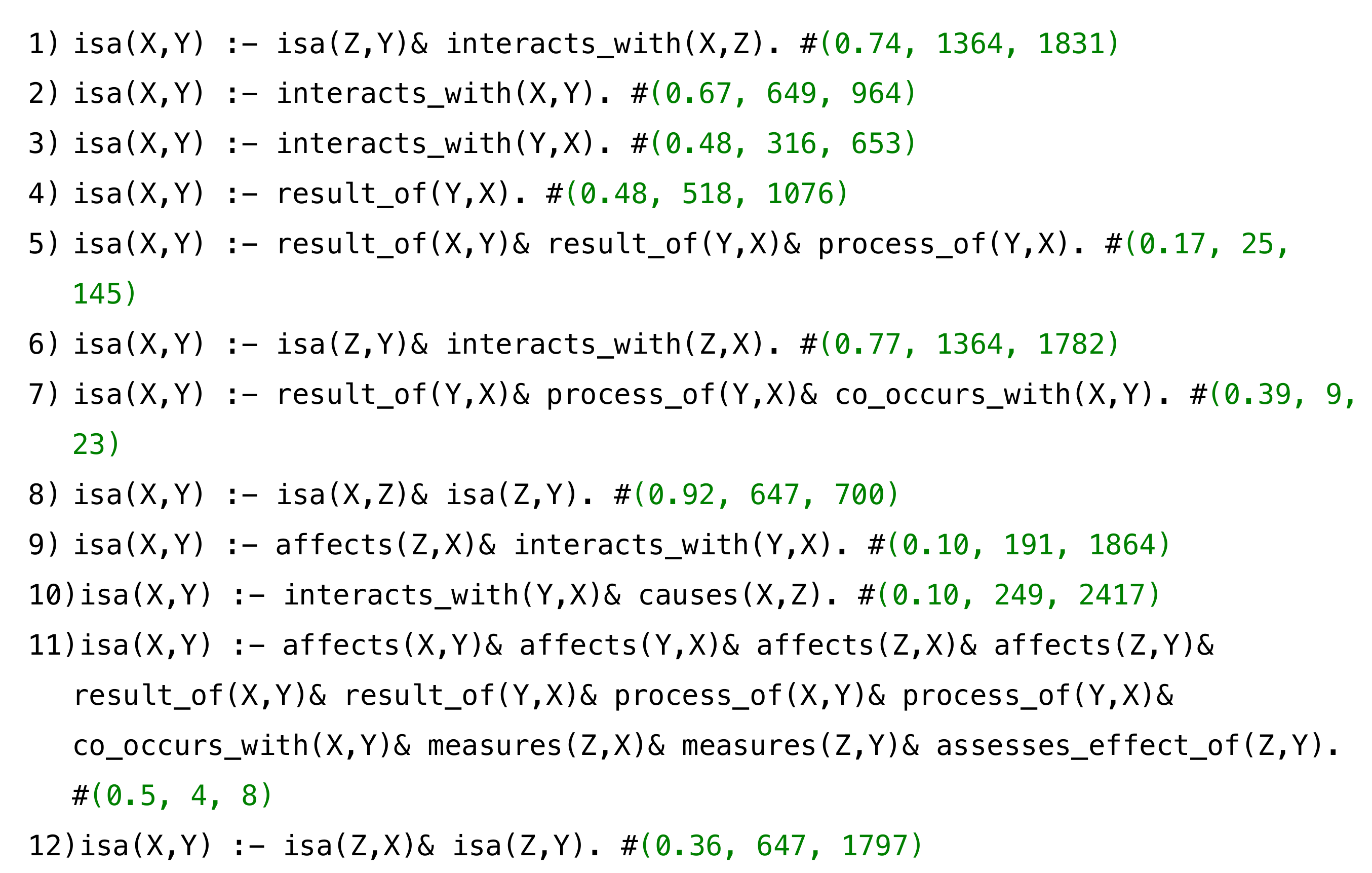}
\end{figure}

\end{enumerate}


\end{document}